\documentclass[letterpaper, 10 pt, journal]{ieeeconf}  

\IEEEoverridecommandlockouts                              

\usepackage{mfirstuc}
\usepackage{graphics} 
\usepackage{epsfig} 
\usepackage{mathptmx} 
\usepackage{times} 
\usepackage{amsmath} 
\usepackage{amssymb}  
\usepackage{siunitx}
\usepackage{verbatim}
\usepackage[noadjust]{cite}
\usepackage{siunitx}
\usepackage{booktabs}
\usepackage{url}
\usepackage{listings}
\usepackage[dvipsnames]{xcolor}
\usepackage[normalem]{ulem}
 \usepackage{ulem} 
 \usepackage{subcaption}
\usepackage[font={small}]{caption}
\usepackage{colortbl}
\usepackage{multirow}
\usepackage{graphicx}
\title{\LARGE \bf Characterization, Experimental Validation and Pilot User Study of the Vibro-Inertial Bionic Enhancement System (VIBES)}

\author{Alessia Silvia Ivani\authorrefmark{4}$^{1,2,3}$, Federica Barontini$^{1,2}$, Manuel G. Catalano$^{1}$,  Giorgio Grioli$^{1,2,3}$,\\  Matteo Bianchi$^{2,3}$, and Antonio Bicchi$^{1,2,3}$   
\thanks{*This article has been accepted for publication in IEEE Transactions on Haptics. This is the author's version which has not been fully edited and
content may change prior to final publication. Citation information: DOI 10.1109/TOH.2024.3435588.
© 2024 IEEE.  Personal use of this material is permitted.  Permission from IEEE must be obtained for all other uses, in any current or future media, including reprinting/republishing this material for advertising or promotional purposes, creating new collective works, for resale or redistribution to servers or lists, or reuse of any copyrighted component of this work in other works.
This work was supported by the European Research Council Synergy Grant Natural BionicS (NBS) project (Grant Agreement No. 810346), by the Italian Ministry of Education and Research (MIUR)  in the framework of the FoReLab project and Crosslab project (Departments of Excellence); by PNRR, M4 C2 I1.5 Ecosistema dell'Innovazione "Tuscany Health Ecosystem (THE)" - Ecosistema dell’innovazione sulle scienze e le tecnologie della vita in Toscana (CUP I53C22000780001) - Spoke 9, and the ERC Proof of Concept Wearable Integrated Soft Haptic Display for Prosthetics (WISH) project (Grant No. 101069179).}
\thanks{$^{1}$with  Soft Robotics for Human Cooperation and Rehabilitation, Istituto Italiano di Tecnologia, Genova 16163, Italy.}{}
\thanks{$^{2}$with Centro di ricerca E. Piaggio, University of Pisa, Pisa 56122, Italy.}
\thanks{$^{3}$ with Department of Information Engineering, University of Pisa, Pisa 56122, Italy.}
 \thanks{\authorrefmark{4} Corresponding author \tt\small alessia.ivani@iit.it}
}

\begin{document}
\maketitle
\thispagestyle{empty}
\pagestyle{empty}

\begin{abstract}
This study presents the characterization and validation of the VIBES, a wearable vibrotactile device that provides high-frequency tactile information embedded in a prosthetic socket.
A psychophysical characterization involving ten able-bodied participants is performed to compute the Just Noticeable Difference (JND) related to the discrimination of vibrotactile cues delivered on the skin in two forearm positions, with the goal of optimising vibrotactile actuator position to maximise perceptual response.
Furthermore, system performance is validated and tested both with ten able-bodied participants and one prosthesis user considering three tasks.
More specifically, in the Active Texture Identification, Slippage and Fragile Object Experiments, we investigate if the VIBES could enhance users' roughness discrimination and manual usability and dexterity. 
Finally, we test the effect of the vibrotactile system on prosthetic embodiment in a Rubber Hand Illusion (RHI) task. 
Results show the system's effectiveness in conveying contact and texture cues, making it a potential tool to restore sensory feedback and enhance the embodiment in prosthetic users.
\end{abstract}
      \vspace{-0.4cm}
\section{Introduction}
In recent years, the pursuit of non-invasive feedback solutions for prosthetic users has gained considerable attention to enhance the sensory experience for individuals with limb loss or impairments. 
Various methods have been explored and investigated to offer a comprehensive and natural feedback system, bridging the gap between the user and their prosthetic device. 
Among these methods, mechano-tactile (force stimuli), electrotactile,  and vibrotactile stimulation stand out as prominent approaches. 
Vibrotactile stimulation has emerged as one of the most extensively investigated techniques, primarily due to its ease in tactile signal modulation and the practical advantages it offers with compact-sized, affordable, and readily available vibrotactile actuators~\cite{choi2012vibrotactile,thomas2019comparison}. 
However, only a limited number of sensorized bionic hands are currently commercially available (e.g. the PSYONIC Ability Hand \cite{akhtar2020touch}, and the Vincent Hand \cite{VincenHand}).
\begin{figure}
  \centering
  \includegraphics[width=0.8\columnwidth]{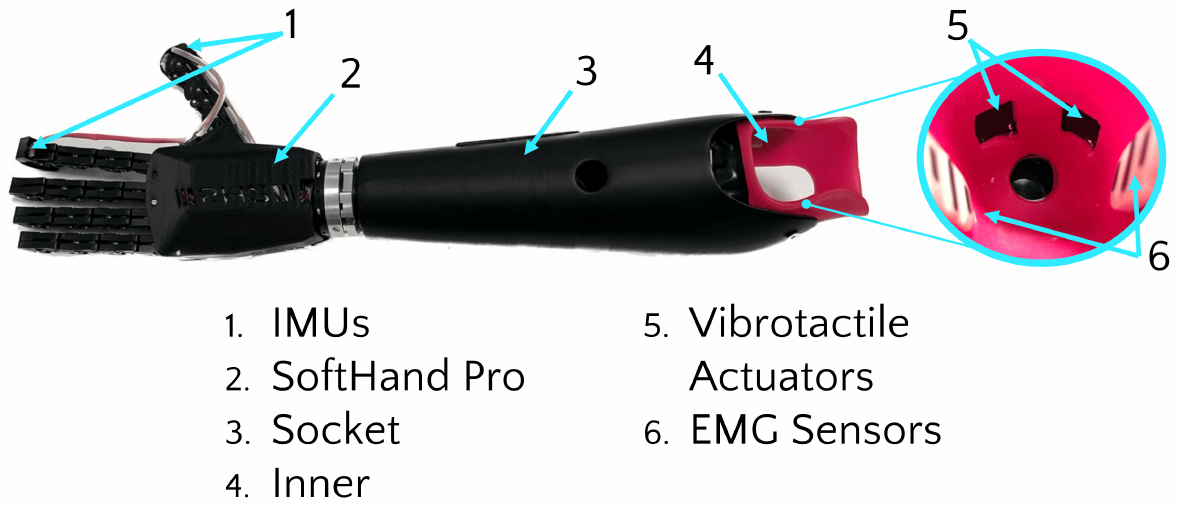}
  \caption{Overview of the main components of the prosthetic device with the VIBES (Vibro-Inertial Bionic Enhancement System). A detailed view (right) shows the inner part of the socket with vibrotactile actuators and EMG sensors.}
  \label{fig:system}
  \vspace{-0.8cm}

\end{figure} 
In the study by Kim and Colgate on grip force control of sEMG-controlled prosthetic hands in targeted reinnervation amputees, the authors emphasize that an optimal and effective haptic feedback system should be integrated into the prosthesis~\cite{kim2012haptic}. The authors argue that this system should incorporate both somatotopic matching (SM) and modality matching (MM) paradigms to achieve a natural and intuitive user experience.
The SM paradigm enables the generation of signals that are transmitted in the place where they would naturally be felt.
Antfolk et al. studied a non-invasive sensory feedback system in twelve trans-radial amputees and twenty healthy non-amputees, providing arm stump sensory feedback for prosthetic hand users \cite{antfolk2012sensory}. They demonstrated that the central nervous system processes stump-level sensory data similarly to finger-level stimuli.
The MM paradigm involves providing a stimulus resembling the original sensation, such as pressure cues. 
Vibrotactile stimulation naturally conveys high-frequency information associated with surface contact, including texture, roughness, and first-contact cues  (Shao et al. \cite{visell}). 
Thus, within the modality matching framework, the integration of vibrotactile stimulation conveys acceleration-mediated contact cues related to the characteristics of touched objects \cite{actionsomatosensory}.

Apart from the challenges associated with the design and development of the system, as emphasized by Sensinger et al. in their review 
, varying perspectives exist on the impact of feedback on prosthesis functionality and performance~\cite{sensinger2020review}.
For instance, in a study by Raveh et al., twelve trans-radial prosthetic users demonstrated significant improvements in a modified Box and Block test with vibrotactile feedback, enhancing time and grasping performances~\cite{raveh2018myoelectric}. Conversely, the study of Markovic et al. involving six prosthetic users in a Box and Block test showed no notable effect of vibrotactile feedback on performance~\cite{markovic2018clinical}.
Therefore, it is paramount to not only design and develop haptic feedback devices but also to assess them rigorously. 
Conducting comprehensive investigations is essential to understand the impact of haptic cutaneous devices and their influence on manual dexterity and usability.

Beyond the scope of functional enhancements, an exploration of user experience, demands attention. 
As technology advances, integrating a prosthetic into a user's body image—embodiment—gains prominence. 
Segil et al. review delves into sensory feedback studies, underscoring the need for further research into the psychological and perceptual dimensions of prosthetic use ~\cite{segil2022measuring}.
The Rubber Hand Illusion (RHI) stands as the most frequently used methodology to study ownership \cite{zbinden2022prosthetic,segil2022measuring}.
Ehrsson et al. extended RHI experiments to amputees, applying tactile stimulation to the residual limb or distal stump  ~\cite{doi:10.3109/02844310903113107}.
D’Alonzo et al. employed a vibrotactile device and brush, yielding high ownership ratings with brush stimulation and slightly lower ratings with vibrotactile stimulation ~\cite{6856175}.
Marasco et al. replaced brush stimulation with a pressure actuator, reporting ownership in reinnervated trans-humeral amputees when visual feedback matched the pressure actuation on their reinnervated skin  ~\cite{marasco2011robotic}. \\
\indent  In our previous work, we presented the Vibro-Inertial Bionic Enhancement System (VIBES) integrated with the SoftHand Pro (SHP),  Godfrey et al. ~\cite{godfrey2018softhand}, a prosthetic hand (Fig.\ref{fig:system}) (Ivani et al. \cite{ivani2023vibes}).
The device comprises two Inertial Measurement Units (IMUs) on the distal phalanx of the index and thumb in the prosthetic hand as sensors and two integrated vibrotactile actuators, transmitting texture and contact cues to specific stump sites. 
To the best of the author's knowledge, the VIBES is considered one of the earliest vibrotactile devices fully incorporated into a prosthetic system for transmitting surface information to the user.
It strives to achieve somatotopic and modality matching paradigms and addresses the absence of intrinsic somatosensory feedback caused by damping elements in the artificial body parts, crucial for precise motor commands  (Amoruso et al.~\cite{amoruso2022intrinsic}). \\
\indent  We conducted physical and preliminary psychophysical characterizations of the VIBES with ten able-bodied subjects, followed by an Active Texture Identification experiment with a prosthetic user (Ivani et al.~\cite{ivani2023vibes}).
In contrast to the preceding study, this manuscript introduces various experiments to extensively evaluate the Vibro-Inertial Bionic Enhancement System and gain a more comprehensive understanding of integrated haptic feedback effectiveness and potential in the field of prosthetics.
The VIBES is here characterized through a psychophysical experiment that utilizes the Just Noticeable Difference (JND) to determine optimal actuator positioning on the skin in two forearm positions (A and B) with the objective of identifying the most sensitive configuration (see Fig. \ref{fig:actcase}).
The Active Texture Identification experiment, aimed at rigorously assessing the effectiveness of the VIBES in providing sensory cues, is conducted with ten able-bodied participants to yield more robust statistical results.
Subsequently, two additional tasks are designed to examine the VIBES' impact in alternative scenarios, particularly in manual dexterity and usability assessments, named Fragile Object and Slippage experiments, wherein the device is actively engaged and potentially capable of influencing performance outcomes. 
These experiments test the functionality of the VIBES with able-bodied subjects and the intended end-user.
With the complete integration of the VIBES, an essential need remains to assess the Rubber Hand Illusion (RHI) experience in the SoftHand Pro prosthesis users.
The RHI experiment is designed to investigate the perception of body ownership, both in the presence and absence of tactile feedback.
This study is essential for deepening our understanding of the connection between sensory augmentation and perceptual integration.
It also guides future developments in devices like the SHP and the VIBES haptic system.
\vspace{-0.3cm}
\section{The VIBES}\label{sec:1}
We report on the description of the system that remains consistent with our prior work (Ivani et al.~\cite{ivani2023vibes}).
The VIBES comprises two planar vibrotactile actuators, an electronic board, and two Inertial Measurement Units (IMUs, MPU-9250) placed on the distal phalanx of the SHP.
\begin{figure}
      \centering
      \includegraphics[width=\columnwidth]{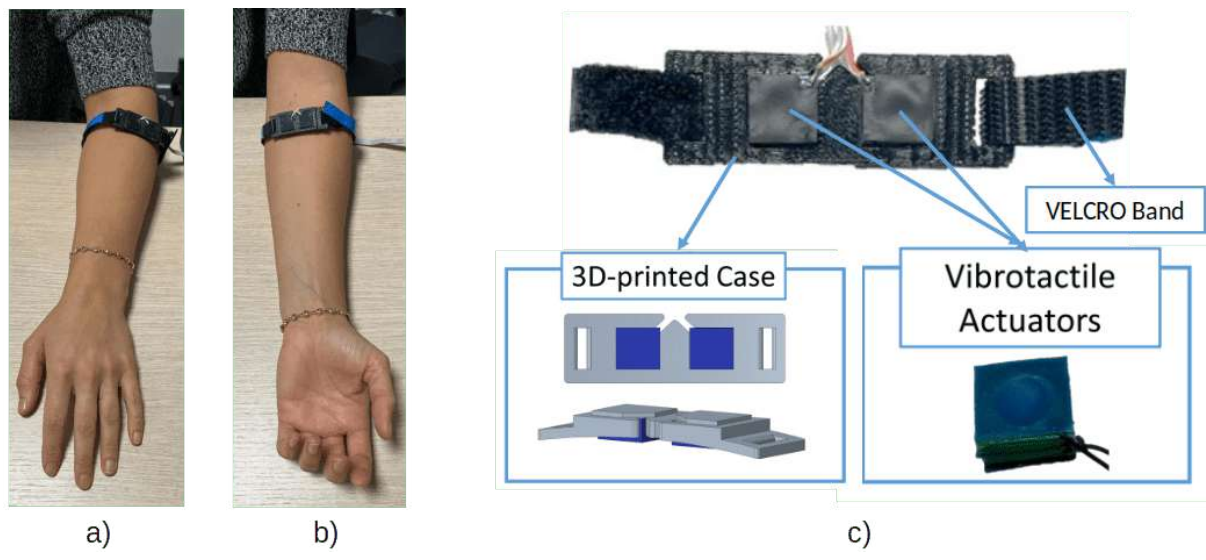}
      \caption{The bracelet worn by an able-bodied subject in the two experimental conditions: A and B (respectively a) and b) in the figure). c) The bracelet of the VIBES for the able-bodied experiments made of a 3D-printed case with vibrotactile actuators and adjustable VELCRO bands for secure fitting.}
      \label{fig:actcase}
      \vspace{-0.6cm}
\end{figure}
The vibrotactile actuator employed is the Haptuator Planar (HP) by TactileLabs\footnote{TactileLabs, Haptuator Planar, Available: \url{http://tactilelabs.com/products/haptics/haptuator-planar/}} (see the highlighted box in Fig.\ref{fig:actcase}c).
The HP is a voice-coil actuator with a 50-500 Hz bandwidth covering the tactile sensitivity range.
Stimuli transmission is perpendicular to the skin, reducing system size compared to tangential transmission.
The 1.8g, 12x12x\SI{6}{mm} HP suits prosthetic integration with its compact shape, soft surface, and skin contact design.

About the control strategy, the recorded acceleration signals $a = (a_x, a_y, a_z)$ from each IMU undergo a sequence of processing steps, detailed in Fani et al. \cite{paperfani}. 
These steps encompass filtering, dimensional reduction, and limiting.
Filtering removes noise from IMUs in the prosthetic system (e.g. free hand motion). 
Subsequently, the three acceleration components are reduced with the DFT321 algorithm \cite{landin2010dimensional}. 
Thus, the three components $a_x, a_y, a_z$ are then reduced into a one-dimensional signal $A$ with the same energy as their sum.
The DTF321, as part of the algorithms that seek the spectral difference, is considered by Lee et al. one of the best 321 approaches for offline processing when perceptual similarity is prioritized~\cite{9773011}.
A scaling factor translates the signals into PWM values to activate the actuators while adhering to the actuators' specifications. The PWM operates at a frequency of 5KHz, with the PWM being updated at 1KHz.
The signal from the index IMU corresponds to the left actuator, whereas the signal from the thumb IMU corresponds to the right.
Real-time control employs electronic boards for signal recording and actuation (Della Santina et al. \cite{NMMI}).
\section{System Characterization}\label{sec:2}
In our prior work, a physical and psychophysical characterization of the system was performed to test the effectiveness of the actuators in conveying reliable texture cues (Ivani et al. \cite{ivani2023vibes}).  
By using the HP actuator on able-bodied participants' index fingers, we determined a Just Noticeable Difference (JND) of 87.30 $\mu$m (95\% CIs: 79.69 - 96.52 $\mu$m), highlighting the actuator's ability to enable accurate perception and discrimination of roughness.

Our current objective is to test how the placement of actuators on the forearm skin influences perception to provide insights for optimizing actuator placement within the prosthetic device.
The experiment assesses participants' ability to differentiate roughness levels conveyed by the actuators on the forearm using the Just Noticeable Difference. 
JND is the smallest detectable change in a stimulus causing a noticeable difference in sensation (Jones et al. ~\cite{jones2012application}). 
The actuator is positioned on participants' left forearm in two distinct experimental conditions, each featuring different skin placement configurations: A and B (please refer to Fig.\ref{fig:actcase}).
The experiment involves a \textit{recording session} and a \textit{user session}, which are analogous to the procedures employed in our prior work \cite{ivani2023vibes} (Section III B).
In this instance, we evaluated the perception of the two experimental conditions through the passive transmission of stimuli to the subjects.

Review No. 30/2020 from the Committee on Bioethics of the University of Pisa granted approval for all experimental procedures reported in this manuscript.
\vspace{-0.4cm}
\subsection{Participants}
Ten able-bodied participants (7 males and 3 females, age mean$\pm$SD: 27,2 $\pm$1,48) are enrolled and give their informed consent to participate in the experiments. 
Participants have no disorder that could affect the experimental outcome.
 A 43-year-old female who experiences agenesis of the left forearm is enrolled to test the system.
The prosthetic user participant exhibits no cognitive impairments that might influence her ability to comprehend or adhere to the study's instructions.
The participant usually wears a cosmetic prosthesis, although she has previous experience with myoelectric prosthetic devices.
\vspace{-0.4cm}
\subsection{Materials}
Five sandpapers sized 28x23 cm are employed for the experimental methods. 
These sandpapers source from RS\footnote{RS, [Online], Available: \url{https://it.rs-online.com/web/}}, adhere to the Federation of European Producers of Abrasive Products (FEPA) P-grading system.
We follow the method of constant stimuli as described by Jones et al. \cite{jones2012application}.
A preliminary study determines the detectable range of sandpapers by the actuator, with grit sizes ranging from 18 $\mu$m (P1000) to 264 $\mu$m (P60).
In selecting stimuli, we follow the procedure outlined by Libouton et al. for a tactile roughness discrimination experiment \cite{libouton2010tactile}. Our choice for the reference stimulus is P120, with an average particle size of 127 $\mu$m, positioned approximately midway between the smoothest and roughest detectable stimuli.
The remaining stimuli are chosen to maintain equal spacing in terms of micrometres and considering the available types of sandpapers. 
Therefore, five stimuli are chosen ranging from the smoothest to the roughest: 18 $\mu$m (P1000), 65 $\mu$m (P220), 127 $\mu$m (P120), 195 $\mu$m (P80), and 264 $\mu$m (P60).

To facilitate the use of the VIBES by participants, a bracelet has been designed for actuator placement (Fig. \ref{fig:actcase}). This bracelet consists of a 3D-printed case for the actuators and a VELCRO band to ensure secure fitting on the forearm.
 
During the \textit{user session}, participants are seated comfortably on an office chair, positioning their forearm or, in the case of a prosthetic user, their residual limb without the prosthesis, on a desk. 
To ensure isolation, participants wear goggles with opaque lenses and headphones that play white noise.
The experiment follows the chosen psychophysical approach via customized C++ software.
\vspace{-0.4cm}
\subsection{Methods}\label{sec:2b}
\subsubsection{Recording Session}
During the \textit{recording session}, tactile stimuli are captured from the IMU of the SoftHand Pro index finger and saved without applying any filtering techniques. Five distinct stimuli are captured as the SoftHand Pro index finger slides across five selected sandpapers. A customized C++ software is developed to capture, segment, and save the accelerations from the IMU on the index fingertip, which records three acceleration components ($a_x$, $a_y$, $a_z$). Boundy-Singer et al. have demonstrated in their research that modifying scanning speeds has minimal impact on texture perception, and alterations in contact force, as highlighted by Saal et al., do not significantly affect texture perception either \cite{boundy2017speed, SAAL201899}.
Therefore, the velocity magnitude and pushing force are considered to have negligible influence on the experimental outcomes.
Still, efforts are made to keep velocity and pushing force relatively constant to ensure consistency in the experiment. 
This is achieved by maintaining a fixed time window for the recordings and ensuring the SHP remains at the same height above the sandpaper.
Two signals are saved for each type of sandpaper. 
These acceleration signals are then mapped into PWM values to activate the haptic actuator, following the control strategy detailed in Section II.
\subsubsection{User Session}
Employing the method of constant stimuli, we compute JND by presenting pairs of stimuli for participants to identify the rougher one  (Jones et al. \cite{jones2012application}). 
The signals recorded during \textit{recording session} are transmitted passively, in a playback modality, onto the user's left forearm (or left residual limb). This entails the utilization of prerecorded signals from the SoftHand index finger with the control strategy described in Section II. We use the HP actuator and the bracelet on the left arm to replicate the prosthetic user scenario (left forearm limb agenesis) while accounting for spatial limitations dictated by EMG sensor placement within the prosthesis.

Only the corresponding left HP actuator is activated since the recorded stimuli originate from the IMU on the SHP index finger. 
In randomized order, able-bodied participants test the actuator in two experimental conditions (A and B).
In each experimental condition, participants experience pairs of stimuli, the reference stimulus, and a comparison stimulus signalled audibly.
After the second stimulus is off, the participant is asked to report which stimulus in the pair felt rougher (“Which of the two stimuli did you perceive to be rougher?”). The subject is instructed to distinguish textures through vibration frequency and intensity on the forearm.
The answer is saved for the subsequent analysis.
Comparison stimuli are transmitted from pre-registered signals following a random permutation of all possible combinations of stimuli, thereby ensuring that all participants encounter the same stimuli. 
The reference and comparison signals are randomly selected from the two saved stimuli per sandpaper type in the \textit{recording session}. 
There is no familiarization or stimulus repetition.
Each experimental condition consists of 100 pairs of stimuli, 20 per stimulus level (across five sandpaper grits).

Based on findings from able-bodied participants, which indicated minimal impact of actuator placement on perception, and to minimize disruption to the prosthetic user, we test the sensitivity of the prosthetic user only in Experimental Condition A, where she anticipates feeling texture cues for the index on the residual limb.
\vspace{-0.3cm}
\subsection{Data Analysis}
Using logistic regression, we explored differences in perceived roughness among the 10 able-bodied participants. Logistic regression assesses the impact of experimental variables on groups through fixed effect parameters exclusively. The model is formulated as follows:
\begin{equation}
\begin{split}
P(Y_j=1) &= \frac{1}{1 + e^{-(\beta_0 + \beta_1*x_j)}},
\label{eq:logit}
\end{split}
\end{equation}
Here, $P(Y_j=1)$ denotes the probability of perceiving the comparison stimulus as rougher than the reference in trial $j$. 
The parameters $\beta_0$ and $\beta_1$ represent fixed effects, capturing the intercept and slope of the linear function (linear predictor) across all subjects. 
The explanatory variable $x_j$ signifies the sandpaper stimuli. Next, for each experimental condition, we derive the Just Noticeable Difference (JND) and the Point of Subjective Equality (PSE) alongside their corresponding 95\% confidence intervals (CIs) with the bootstrap method. 
JND and PSE are computed as $\frac{1}{|\beta_1|}$, where $|\beta_1|$ denotes the absolute value of the coefficient and  $-\frac{\beta_0}{\beta}$, respectively \cite{10.1167/12.11.26}.
Then, to test the normality of data we use a Shapiro Wilk test on JND and PSE values. Thus, we use a Wilcoxon test to verify if there is any significant difference between the two experimental conditions.
\begin{figure}
      \centering
      \includegraphics[width=0.9\columnwidth]{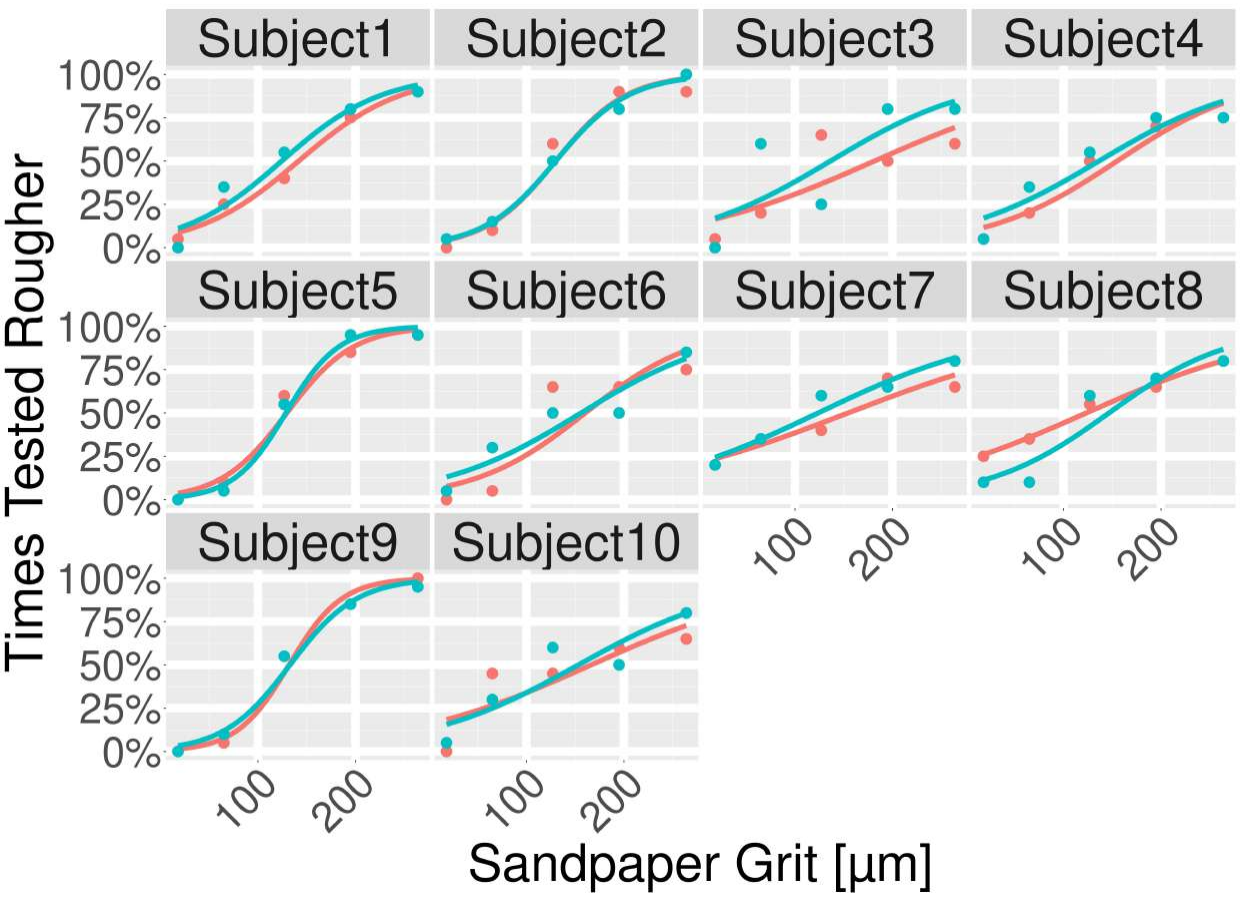}
      \caption{Logistic fit for the ten subjects. Raw data and model predictions for each participant are labelled as 1 to 10. Blue: A condition; red: B condition.}
      \label{fig:jndhs}
       \vspace{-0.6cm}
\end{figure}
Since only one prosthetic user was involved, a probit fit on results was performed.
\vspace{-0.4cm}
\subsection{Results}
\textbf{Able-bodied} - The logistic fitting to the data is illustrated in Fig. \ref{fig:jndhs}.
We evaluated the capacity of participants to discriminate roughness stimuli rendered by the HP actuator on the forearm in two experimental conditions: A  and B condition.
We modelled 10 subjects' data as in (\ref{eq:logit}). 
Confidence intervals of the JND and PSE in the two experimental conditions were overlapping.
The JND was equal to 58.79 $\mu$m (95\% CIs:  54.41- 66.19 $\mu$m) for the A experimental condition and to 64.10 $\mu$m (95\% CIs:  58.19 - 69.42 $\mu$m) for the B experimental condition.
The PSE was equal to 136.73 $\mu$m (95\% CIs: 127.79 - 146.57  $\mu$m) for the A experimental condition and to  145.78 $\mu$m (95\% CIs: 135.62 -156.08 $\mu$m) for the B experimental condition.
From the Shapiro–Wilk test, the JND and PSE data were not normally distributed thus, we performed a Wilcoxon test to test significant differences. 
From the Wilcoxon test, the two experimental conditions were not significantly different (p$>$0.05 for both JND and PSE values).
The participants effectively differentiated between the diverse textures, and their responses remained unaffected by the two experimental conditions.
\textbf{Prosthetic User} - Results show the subject achieved 100\% accuracy in distinguishing the roughest (1) and smoothest (5) stimuli from the reference (3), and 95\% accuracy for stimulus (4). However, accuracy dropped to 50\% for distinguishing stimulus (2) from the reference. From a probit fit on the data, the JND was 44.07 $\mu$m.
\vspace{-0.4cm}
\section{System Validation and Pilot Experiments}\label{sec:3}
Experiments are designed to validate the VIBES with able-bodied participants and to test the system with a prosthetic user.
Our experiments aim to comprehensively assess the effectiveness of the VIBES when integrated with the SoftHand Pro for texture discrimination, which is its primary function.
Additionally, two supplementary tasks evaluate the VIBES influence on manual dexterity and usability.
Three distinct tasks are the Active Texture Experiment, the Fragile Object Experiment, and the Slippage Experiment.
The experiments test the effectiveness of the feedback delivered by the VIBES device in conjunction with the SoftHand Pro under controlled conditions: these conditions include simulating low-light scenarios and allowing users to focus on the feedback~\cite{raveh2018myoelectric}.

At the end of the three experiments, all subjects fill out a NASA Raw Task Load Index (NASA RTLX) questionnaire to evaluate the workload during each task with and without the VIBES (Georgsson et al.\cite{Georgsson2020NASARA}).
Participants are prompted to evaluate their level of agreement on a seven-point Likert-scale survey.
This evaluation encompasses questions about the system's usability, comfort, and performance during the experimental tasks. 
Additionally, the prosthetic user participant assesses the system usability using the System Usability Scale (SUS) (Lewis at al. ~\cite{Lewis2018TheSU}).

A RHI Experiment is also conducted to evaluate SHP ownership of a prosthetic user. Segil et al. introduces a three-domain embodiment concept, comprising ownership, body representation, and agency \cite{segil2022measuring}.
Ownership is the belief that a limb or tool is part of oneself, expressed as "part of my body", involving the recognition of the limb, tool, or device as an integral part of oneself rather than an external entity.
Ownership illusions, such as the rubber hand illusion (RHI), allow researchers to investigate the neurobiological and perceptual processes underlying these experiences.
We have adapted the protocol from Marasco et al. for testing the ownership of the SoftHand Pro (SHP) with the VIBES device in a prosthetic user \cite{marasco2011robotic}.
\vspace{-0.4cm}
\subsection{Participants}
Twelve able-bodied participants (9 males and 3 females, age mean$\pm$SD: 28$\pm$2,3) are enrolled and give their informed consent to participate in the Active Texture Experiment.
Ten able-bodied participants (7 males and 3 females, age mean$\pm$SD: 27.2$\pm$1,3) are enrolled and give their informed consent to participate in the Fragile Object Experiment and the Slippage Experiment; eight of them also perform the Active Texture Experiment.
Participants are chosen from individuals employed in the same research institute, with the only exclusion criterion being any physical or psychological impairments that could potentially affect the outcomes of the experiment, provided they are unfamiliar with haptic feedback devices and have no prior experience with vibrotactile feedback.
The same prosthetic user participant described in Section III A is involved in the pilot experiments.
\vspace{-0.4cm}
\subsection{Materials}
We integrate the VIBES system into a SHP for the prosthetic user and design a handle for able-bodied participants.
For the SHP, the same control strategy as in \cite{barontini2021wearable} is used.
The electrical activity of the Flexor Digitorum Superficialis (FDS) and Extensor Digitorum Communis (EDC) muscles is measured on the user's forearm skin using two Ottobock 13E200 sensors\footnote{OttoBock HealthCare GmbH, http://www.ottobock.com/}. 
These sensors provide amplified, bandpass-filtered, and rectified versions of the raw sEMG signals.
  
When integrating the VIBES system into the prostheses, the placement of actuators is determined by spatial limitations imposed by the positioning of EMG sensors and the confined space within the prosthetic device.
Additionally, building upon the findings presented in Section II, which indicated that actuator placement had minimal impact on perception, we position the actuators in alignment with the sensitivity of the prosthetic user, where she anticipates feeling texture cues for the index and thumb on the residual limb.
As a result, we position both actuators on the A experimental condition of the residual limb, with one situated laterally corresponding to the index IMU sensor and the other placed medially for the thumb. 
This arrangement is chosen to effectively address the limited space available and ensure the optimal positioning of the actuators within the specified configuration.
Please refer to Fig. \ref{fig:system}.
This configuration is adopted also for able-bodied participants. 

A 3D-printed handle is designed for able-bodied participants to facilitate the use of the SHP with their left hand (see Fig.\ref{fig:expab} a).
The handle provides support for positioning the left hand accurately over the SHP and incorporates a VELCRO band to ensure secure attachment.
A quick disconnect wrist mechanism links the robotic hand to the handle.
The SHP's electronic board is connected to the EMG sensor, HP actuators and a battery via cables. 
EMG sensors and the bracelet with the actuators are positioned on the left forearm of able-bodied participant.

Before each experiment, a training phase, which spans 10 minutes, is carried out with all the participants to familiarize them with the control of the SHP through the EMG sensors.
This phase involves the participants grasping and moving various objects, such as a ball or a mug.
Before each experiment starts, the participant sits on a chair near a table.
The subject is isolated in the main experimental phase using white noise and obscured lenses.
\vspace{-0.4cm}
\subsection{Methods}
\begin{figure}
  \centering
   \includegraphics[width=\columnwidth]{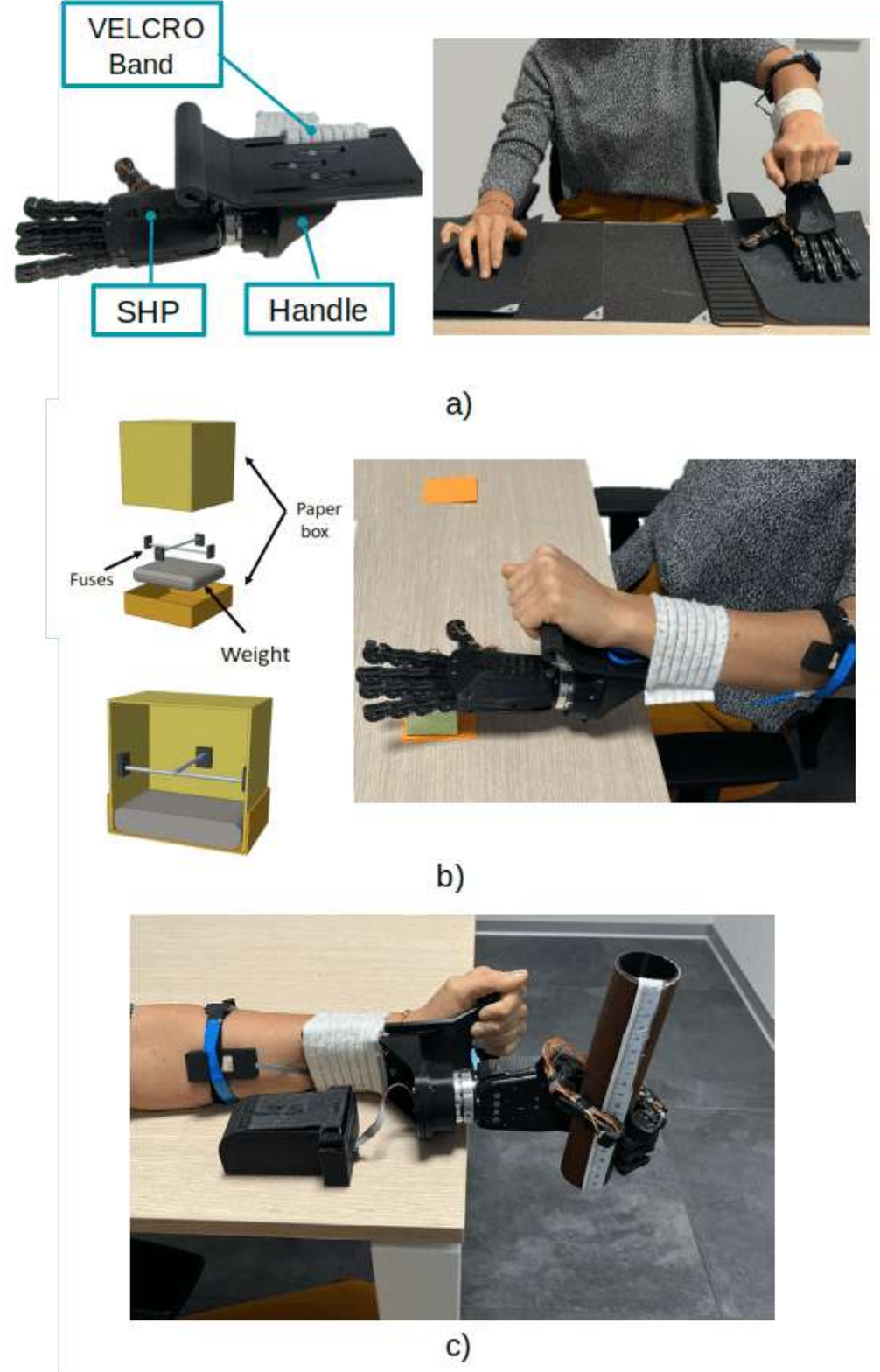}
  \caption{System Validation - An able-bodied user during: a) the Active Texture Identification Experiment in which the subject matches the sandpaper beneath the SHP with the handle (detailed view on the right) by exploring five options with their real hand on the right side; b) the Fragile Object experiment in which the subject moves a fragile object with a detailed view of it; c) the Slippage Experiment in which the subject grasps a cylinder and tries to detect slippage.}
  \label{fig:expab}
  \vspace{-0.8cm}
\end{figure}

\begin{figure}  
\centering
  \includegraphics[width=0.85\columnwidth]{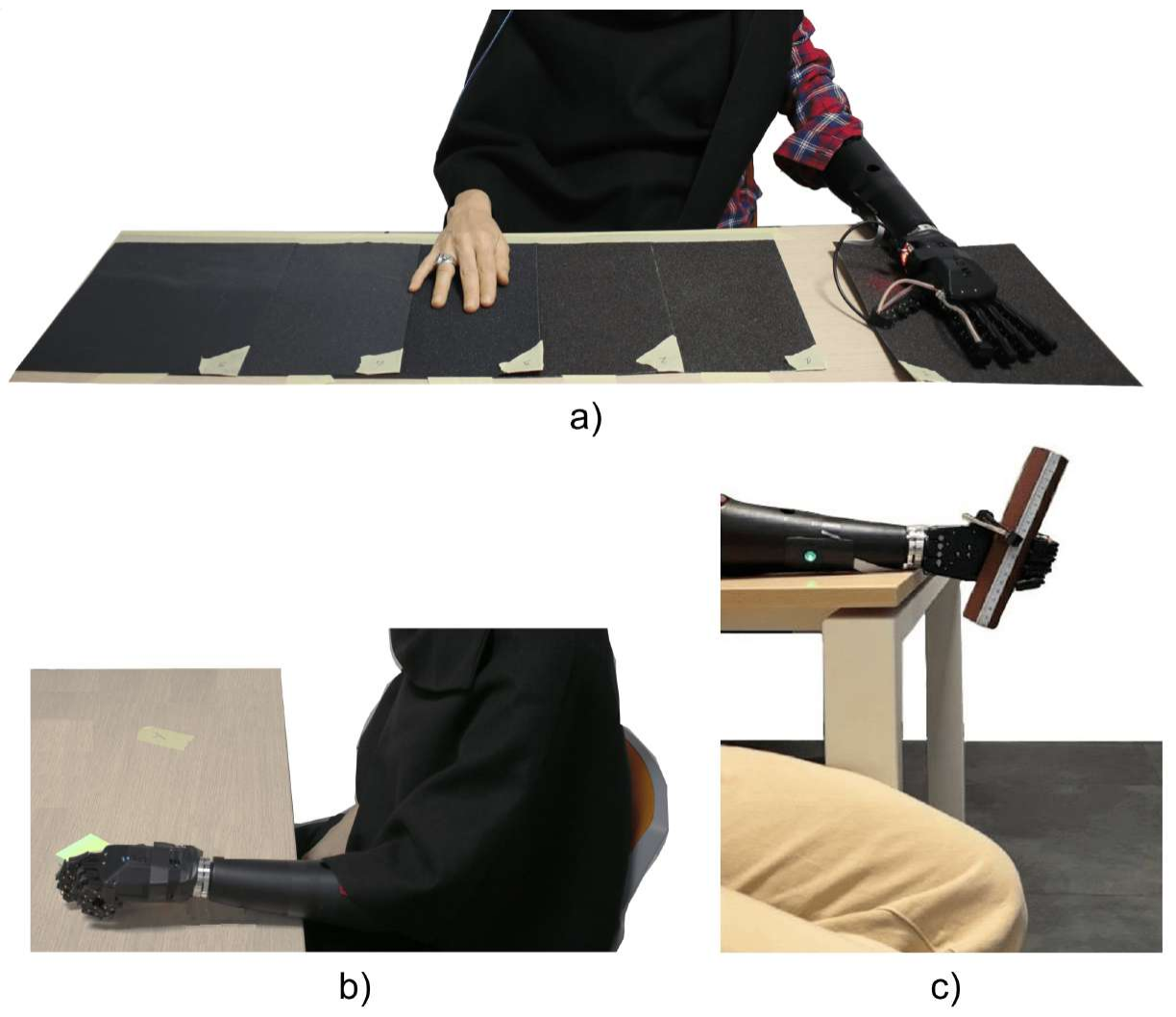}\quad
  \caption{Pilot Experiments - The prosthetic user during: a) the Active Texture Identification Experiment in which the subject matches the sandpaper beneath the SHP by exploring five options with her real hand on the right side; b) the Fragile Object experiment in which the subject move a delicate object; c) the Slippage Experiment in which the subject grasps a cylinder and tries to detect slippage.}
  \label{fig:expsetup}
  \vspace{-0.8cm}
  \end{figure}

In the following, validation, pilot tasks and questionnaires are outlined. Fig. \ref{fig:expab} and Fig. \ref{fig:expsetup} illustrate the setup and the experiment with an able-bodied and a prosthetic user, respectively.
\subsubsection{Active Texture Identification Experiment}
In the initial 10-minutes familiarization phase, the participant can explore a P150 sandpaper (which differs from the one used in the experimental phase) on the table with the SHP. 
The VIBES is turned on throughout this phase, and the subject is not isolated.
Arranged in descending order of roughness, numbered from 1 (P60) to 5 (P1000), the identical set of five sandpapers as those used in Section \ref{sec:2} is positioned on the right side of the table.
The participant explores this set using their right hand.
During each trial, the subject is given one sandpaper beneath the left prosthetic hand, allowing them to explore it using the handle or the prosthesis.
The participant's goal is to determine the corresponding sandpaper from the set of five choices on the right side.
The sandpapers are randomly presented five times each, resulting in a total of 25 identification attempts in two experimental conditions: one with VIBES feedback active and the other without.
\subsubsection{Fragile Object Experiment}
The participants undertake a task similar to that described by Engels et al. to evaluate grasping efficiency and by Barontini et al. for our force feedback device, the WISH device\cite{Engels2018DigitalEW,barontini2021wearable}. The  Wearable Integrated Soft Haptics device (WISH), utilizes two soft pneumatic actuators in contact with the subject's skin to simulate hand grasping force.
In this experiment, each participant is tasked with transferring a fragile object from point \textit{X} to point \textit{Y} ten times without breaking it. The fragile object is a paper box measuring 50x50 mm and weighing 100 grams, housing a fragile fuse positioned at its centre. This fragile fuse is a thin piece of pasta (a spaghetto), susceptible to breaking under excessive pressure from a tight pinch grasp (Fig. \ref{fig:expab}b). Throughout the experiment, participants are instructed to complete the task as fast as possible, though there is no imposed time constraint. Each experiment iteration involves conducting the task accompanied by white noise and partially obscured goggles, both with and without the VIBES system. The metrics recorded for each trial are the time taken to relocate the object from point \textit{X} to \textit{Y} (the timing starts upon the subject's initial contact with the fragile object and stops upon box release of the box), how many fuse breakage, and regrips.
\subsubsection{Slippage Experiment}
During the experiment, each participant lays their left arm or prosthesis on a table.
The handle or the prosthesis are positioned in such a way that the SHP wrist is resting on the table while the SHP itself is suspended in the air.
The subject is isolated and has to grip a graduated cylinder with the SHP. 
The cylinder is covered with a p150 sandpaper to favour grip, and it is 20 cm long.  
The experimenter verifies the consistency of hand grip across trials and subjects by ensuring that each participant grasps the cylinder in the middle.
Fig.\ref{fig:expab}c and Fig.\ref{fig:expsetup}c show the setup of the experiment. 
When the experiment starts, the experimenter disturbs the cylinder grasp with random tactile cues on the cylinder.
Then, the experimenter slightly opens the SoftHand with the PC randomly after twenty seconds from the experiment start, and the cylinder begins to slip.
The subject is asked to detect the slippage from the random tactile cues and regrip the cylinder before it falls using the EMG sensors without moving the handle or the prosthesis from the table.
The cylinder has a graduated scale to measure the amount of slippage (cm). 
The experiment is video recorded to evaluate the reaction time and slip.
The experiment is repeated five times for each experimental condition (with the VIBES and without it). 
The amount of slip is computed from the graduated cylinder, given a reference point on the robotic hand:
\begin{equation}
    Slip= x_{start} - x_{finish},
\end{equation}
Where $x_{start}$ is the point at which the SHP initially grasps the cylinder, and $x_{finish}$ is the point in which the SHP grasps the cylinder after slippage. 
If the cylinder falls, the slip is computed considering only the starting point of the grasp.
The reaction time is computed by analyzing the video recorded and computed as the difference between the time at which the SHP starts to open by the experimenter control and the time at which the SHP starts to close by the subject control.
\subsubsection{Questionnaires}
Two questionnaires are administered to the subjects following the experiments.
The NASA RTLX evaluates the workload through six questions rated with a score from 0 to 20 \cite{Georgsson2020NASARA}:
\begin{itemize}
\item ”How mentally demanding was the task?” (mental demand);
\item ”How physically demanding was the task?” (physical demand); 
\item ”How hurried or rushed was the pace of the task?” (temporal demand);
\item ”How successful were you in accomplishing what you were asked to do?” (performance);
\item ”How hard did you have to work to accomplish your level of performance?” (effort); 
\item ”How insecure, discouraged, irritated, stressed, and annoyed were you?” (frustration). \\
\vspace{-0.3cm}
\end{itemize}
A 0 rate represents very low, while 20 signifies very high, except for the performance in which 0 means perfect performance, and 20 means failure.
Table \ref{tab:likert} shows the 16 questions of the qualitative questionnaire evaluated on a 7-point Likert scale, ranging from 1 (totally agree) to 7 (totally disagree). 
\subsubsection{Rubber Hand Illusion Experiment}
 
The RHI is generated on the left arm of the subject.
The participant watches the experimenter randomly touching the SHP thumb and index fingers IMUs while the HP actuators on the residual limb vibrate accordingly. 
The participant sits on a chair in a quiet room and does not wear the prosthesis.
The actuators are positioned on the residual limb with the bracelet used for the able-bodied participants.
The SHP is positioned in an anatomically correct position on a table in front of the participant.
The shoulder of the user and the proximal end of the SHP on the table are covered with a cloth. 
The RHI is assessed on the left residual arm in four experimental conditions:  
\begin{itemize}
    \item Synchronous Feedback (SF): the HP actuators vibrate synchronously to the acceleration detected by the IMUs on the index and thumb fingernail, and the subject watches the experimenter randomly touching the IMUs on the SHP;
    \item Asynchronous Feedback (AF): the HP actuators vibrate with 500 ms of delay with respect to the acceleration recorded from the IMUs on the index and thumb fingers. The subject watches the experimenter randomly touching the SHP and the IMUs on the SHP. A 500 ms delay is chosen since previous studies have shown that this duration effectively reduces illusory experiences when implementing visuotactile feedback \cite{Kokkinara2014MeasuringTE,10.1371/journal.pone.0087013,articlegrasp};
    \item Visual Only (VO): the experimenter randomly touches the IMUs on the SHP while the HP actuators are not active.
    \item Fixation (F): the SHP is covered, and the HP actuators are not active. The subject's gaze is fixed on a point on the table positioned just beyond the prosthetic hand.
\end{itemize}
As a control, the RHI is also generated in a fifth experimental condition, Contralateral (CL), with the contralateral intact limb in the RHI traditional protocol with the right arm hidden inside a box and a right SHP in place of the hidden hand.
In this case, the experimenter randomly touches the real right thumb and index fingers and simultaneously the corresponding thumb and index SHP fingers.
The AF, VO, F, and CL are control experimental conditions.
The experimental conditions are presented randomly to the subject in three trials for 180 seconds each (Huynh et al. \cite{articlegrasp}).
A 1-minute interval of inactivity, during which the subject fixates on a mark on the table, is provided before and after each trial.
Following each trial, the subject completes a subjective questionnaire containing nine statements.
Each statement can be assessed using a seven-point visual analogue scale, which spans from $'$strongly disagree$'$  $(---)$ to $'$strongly agree$'$ ($+++$).
The nine statements are reported in Table \ref{tab:quest}.
\begin{table*}[]
\small
\MFUnocap{for}%
\MFUnocap{the}%
\MFUnocap{of}%
\MFUnocap{and}%
\MFUnocap{in}%
\MFUnocap{from}%
\MFUnocap{with}%
\caption{RHI: List of the Nine Statements of the RHI Subjective Questionnaire.}
\centering
\label{tab:quest}
\resizebox{\textwidth}{!}{%
\begin{tabular}{lc}
\rowcolor[HTML]{EFEFEF} 
\multicolumn{2}{c}{\cellcolor[HTML]{EFEFEF}\textbf{Statements}} \\ \hline
\rowcolor[HTML]{D3ECC9} 
\multicolumn{1}{l|}{\cellcolor[HTML]{D3ECC9}(1) I felt the touch of the investigator on the prosthetic hand.} &
  \cellcolor[HTML]{D3ECC9} \\
\rowcolor[HTML]{D3ECC9} 
\multicolumn{1}{l|}{\cellcolor[HTML]{D3ECC9}(2) It seemed as if the investigator caused the touch sensations that I was experiencing.} &
  \cellcolor[HTML]{D3ECC9} \\
\rowcolor[HTML]{D3ECC9} 
\multicolumn{1}{l|}{\cellcolor[HTML]{D3ECC9}(3) It felt as if the prosthetic hand was my hand.} &
  \multirow{-3}{*}{\cellcolor[HTML]{D3ECC9}Statements for ownership} \\
\rowcolor[HTML]{F6F0D8} 
\multicolumn{1}{l|}{\cellcolor[HTML]{F6F0D8}(4) It felt as if my residual limb was moving towards the prosthetic hand.} &
  \cellcolor[HTML]{F6F0D8} \\
\rowcolor[HTML]{F6F0D8} 
\multicolumn{1}{l|}{\cellcolor[HTML]{F6F0D8}(5) It felt as if I had three arms.} &
  \cellcolor[HTML]{F6F0D8} \\
\rowcolor[HTML]{F6F0D8} 
\multicolumn{1}{l|}{\cellcolor[HTML]{F6F0D8}(6) I could sense the touch of the investigator somewhere between my residual limb and the prosthetic hand.} &
  \cellcolor[HTML]{F6F0D8} \\
\rowcolor[HTML]{F6F0D8} 
\multicolumn{1}{l|}{\cellcolor[HTML]{F6F0D8}(7) My residual limb began to feel rubbery.} &
  \cellcolor[HTML]{F6F0D8} \\
\rowcolor[HTML]{F6F0D8} 
\multicolumn{1}{l|}{\cellcolor[HTML]{F6F0D8}(8) It was almost as if I could see the prosthesis moving towards my residual limb.} &
  \cellcolor[HTML]{F6F0D8} \\
\rowcolor[HTML]{F6F0D8} 
\multicolumn{1}{l|}{\cellcolor[HTML]{F6F0D8}(9) The prosthesis started to change shape, colour and appearance so that it started to (visually) resemble the residual limb.} &
  \multirow{-6}{*}{\cellcolor[HTML]{F6F0D8}Control statements}
\end{tabular}%
}
\vspace{-0.4cm}
\end{table*}{}
The initial three statements evaluate ownership, while the remaining six are included as controls to assess task compliance and suggestions (Marasco et al. \cite{marasco2011robotic}).
\vspace{-0.4cm}
\subsection{Data Analysis}
\subsubsection{Active Texture Identification Experiment}
Confusion matrices and accuracy metrics, as outlined in the work by Grandini et al.~\cite{grandini2020metrics}, are used to assess the subject's performance.
A Friedman test is used to compare the two modalities.
\subsubsection{Fragile Object Experiment}
The average reaction time, the average number of regrips per participant, and the total number of broken objects of the participants are computed.
A Friedman test is used to compare execution time, broken objects, and the number of regrips with and without the VIBES.
\subsubsection{Slippage Experiment}
The average reaction time, the average slip per person, and the total number of fallen cylinders are computed.
A Friedman test is used to compare the number of fallen cylinders, the reaction time and the amount of slip with and without the VIBES.
\subsubsection{Questionnaires}
We compute the NASA RTLX index for each experiment by transforming scores to a 0 to 100 scale, calculating the mean, and identifying the average among participants. 
Furthermore, we employ the Lilliefors test to assess the normal distribution of NASA RTLX scores. 
Consequently, we perform a paired two-sample t-test to compare the results with and without dual feedback of the NASA RTLX scores.

About the qualitative questionnaire, we compute the mean and the standard deviation of the scores of the ten participants. 
\subsubsection{Rubber Hand Illusion Experiment}
The RHI results are analyzed based on Marasco et al. analysis\cite{marasco2011robotic}.
The numerical values for each question are averaged across three trials of each experimental condition. 
To calculate 95\% confidence intervals (CIs), we use a multiple comparisons procedure with Tukey's honestly significant difference criterion, implemented in Matlab (MathWorks Inc., Natick, USA). 
Errors are pooled across all the experimental conditions and questions to have conservative CIs.
Statements are considered significantly different if their CIs do not overlap \cite{marasco2011robotic}.
\vspace{-0.4cm}
\subsection{Results}
\subsubsection{Active Texture Identification}
\begin{figure}
\centering
\includegraphics[width=0.85\columnwidth]{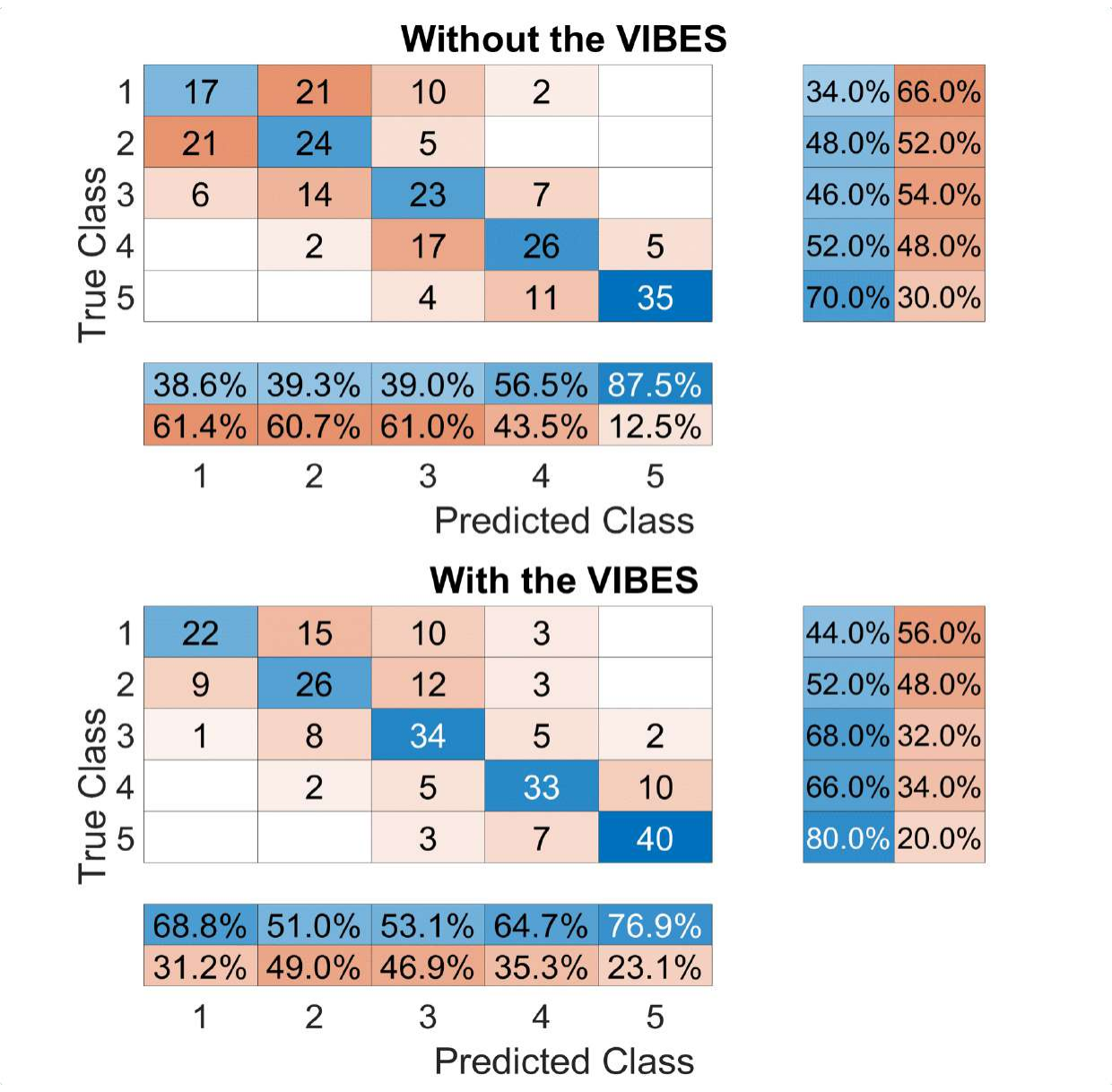}\quad
\caption{System Validation - Active Identification Experiment: confusion matrices with and without the VIBES. Stimuli in descending roughness order, from 1 (P60) to 5 (P1000) (please refer to Section IIIB). The row and column summary displays the percentage of correctly classified and incorrectly classified observations for each true or predicted class.}
\label{fig:active}
\vspace{-0.6cm}
\end{figure}

\textbf{Able-bodied} - About able-bodied participants, two outliers out of twelve were identified, exhibiting low sensitivity to texture on the forearm (i.e. the subjects did not perceive significant stimuli on the forearm), resulting in an exclusion rate of 17\%.
Results of ten subjects are presented through two confusion matrices reported in Fig. \ref{fig:active}, one for each experimental condition, with the relative discrimination accuracy for each sandpaper. 
The correct answers of all tests are reported in the diagonal. 
The confusion matrices show an improvement in sandpaper matching when the subject wears the feedback device compared to without feedback.
The average relative accuracy without the VIBES was 50\% (SD 15\%), while the average relative accuracy with the VIBES was 62\% (SD 9\%). 
The chance level was 20\%.
A Friedman test was used to compare the two modalities, considering 5 trials for each sandpaper type.  
A significant difference was found (p $<$ 0.05) between the two conditions.
Thus, the performance in actively identifying the matching sandpaper of the 10 subjects with the VIBES is significantly better than without it.
We performed Wilcoxon signed rank tests between the scores obtained with the VIBES and those obtained without it, referred to each sandpaper type.
In post-hoc analysis, we compared the scores of each sandpaper type using a two-tailed Wilcoxon signed-rank test with false discovery rate (FDR) adjustment through the Benjamini-Yekuteli correction.
The analyses revealed statistically significant differences between texture 1 and texture 5, and texture 2 and texture 5 for both experimental modalities.
Significant differences were also found between texture 3 and texture 5 and texture 4 and texture 5 in the without the VIBES experimental condition, and texture 1 and texture 3 and texture 1 and texture 4 in the with the VIBES experimental condition. 

\begin{figure}
\centering
\includegraphics[width=0.82\columnwidth]{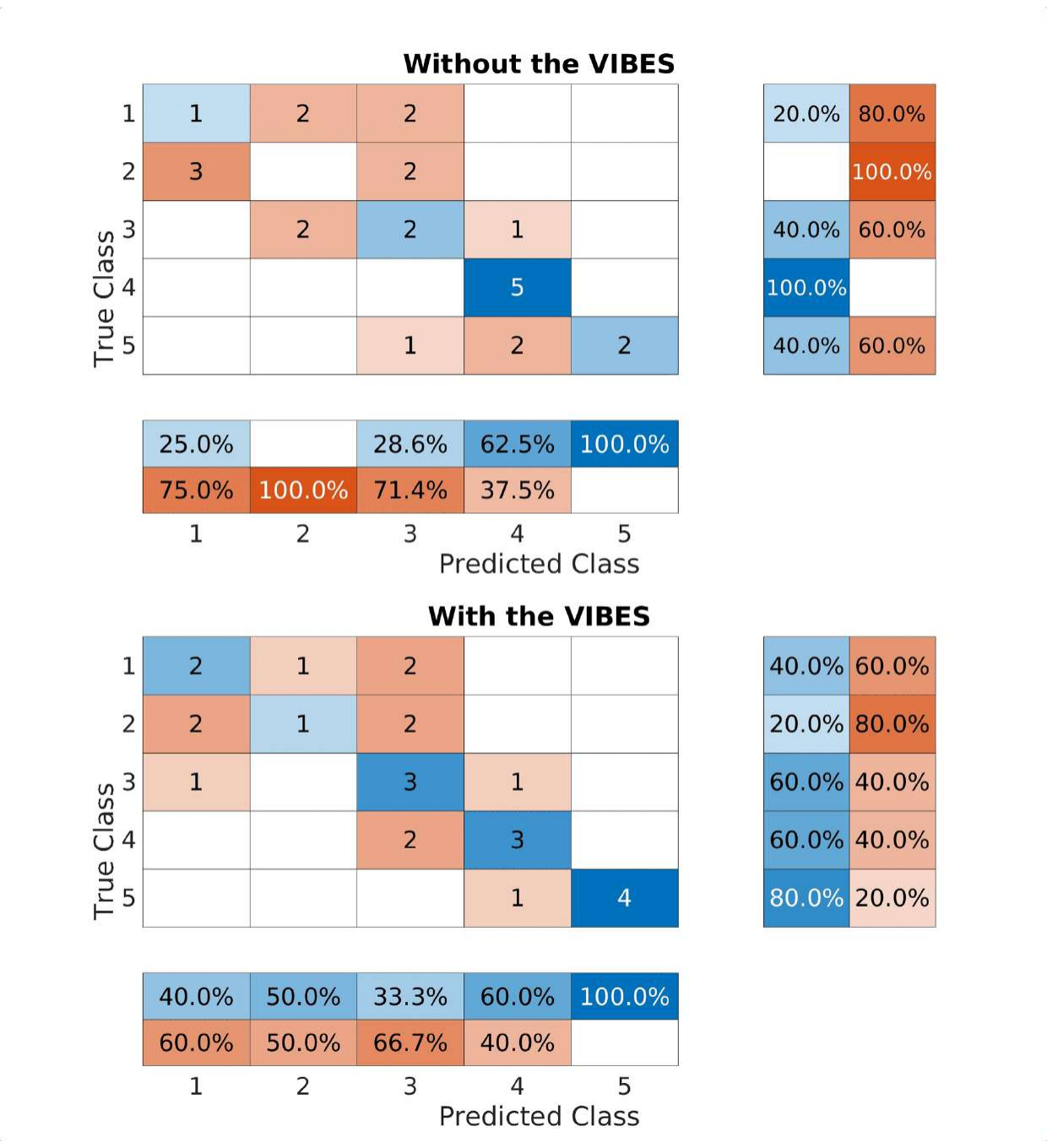}\quad
\caption{Pilot Experiments - Active Identification Experiment: Confusion matrices with and without the VIBES. Stimuli in descending roughness order, from 1 (P60) to 5 (P1000) (please refer to Section IIIB). The row and column summary displays the percentage of correctly classified and incorrectly classified observations for each true or predicted class.}
\label{fig:ati}
\vspace{-0.3cm}
\end{figure}

\textbf{Prosthetic User} - The results of the Active Texture Identification experiments of the prosthetic user are shown in Fig.\ref{fig:ati}.
The confusion matrices show a major improvement in sandpaper identification when the subject wears the feedback device compared to without feedback.
The overall accuracy with the VIBES was 52\%, and without the VIBES was 40\%.
\subsubsection{Fragile Object Experiment}
\textbf{Able-bodied} - With the VIBES, the average time to move the fragile object was 3.7 s, and the number of broken fuses was 41 with 0.46 average regrips. 
Without the VIBES, the average time to move the fragile object was 3.8 s, and the number of broken fuses was 41 with 0.7 average regrips. 
Thus, the number of broken fuses was the same with and without the VIBES. 
The average time to move the object and the average regrips were slightly lower with the VIBES compared to without it.
From the Friedman tests, no statistically significant difference (p$>$0.05) was found between the use of the VIBES feedback and without the VIBES for all experimental outcomes measures (time to move the fragile object, number of broken fuses and number of regrips).

\textbf{Prosthetic User} - About the prosthetic user, Table \ref{tab:isolated} shows the results of the Fragile Object experiment. 
The feedback resulted in a decrease in the total time needed for the ten trials, from 74.1s to 54.8s, and an increase in the number of broken eggs, from 3 to 6.
Regarding the regrips, 2 were detected without the VIBES and 1 with the VIBES.
\begin{table}[]
\small
\MFUnocap{for}%
\MFUnocap{the}%
\MFUnocap{of}%
\MFUnocap{and}%
\MFUnocap{in}%
\MFUnocap{from}%
\caption{\capitalisewords{Pilot Experiments - Fragile Object Experiment: Time to move the fragile objects and number of broken eggs with and without the feedback system.}}
\label{tab:isolated}
\resizebox{\columnwidth}{!}{
\begin{tabular}{c|cc|cc}%
\multirow{2}{*}{\textbf{Trials}} & \multicolumn{2}{c|}{\textbf{Without VIBES}} & \multicolumn{2}{c}{\textbf{With VIBES}} \\ 
                                 & Time to move {[}s{]}     & Broken Fuse     & Time to move {[}s{]}    & Broken Fuse   \\ \hline
1              & 13.0 & 1 & 5.3  & 1 \\
2              & 7.9  & 0 & 5.5  & 1 \\
3              & 6.5  & 0 & 7.0  & 1 \\
4              & 6.2   & 0 & 4.9  & 0 \\
5              & 10.7 & 1 & 6.8  & 0 \\
6              & 7.5  & 1 & 5.6  & 1 \\
7              & 5.3   & 0 & 3.0  & 0 \\
8              & 4.0  & 0 & 10.1 & 1 \\
9              & 5.2   & 0 & 3.1  & 1 \\
10             & 7.8  & 0 & 3.5   & 0 \\ \hline
\textbf{Total} & 74.1 & 3 & 54.8 & 6
\end{tabular}%
}
\vspace{-0.6cm}
\end{table}
\subsubsection{Slippage Experiment}
\textbf{Able-bodied} - The experimental outcome measures were the number of fallen cylinders, the amount of slip, and the reaction time.
With the VIBES, the average reaction time was 0.55 s, with 29 fallen cylinders and an average slip of 7.7 cm. 
Without the VIBES, the average reaction time was 0.7s, with 33 fallen cylinders and an average slip of 7.6 cm.
The number of fallen cylinders and the reaction time were lower when the VIBES was active.
The average slip was slightly lower without the VIBES.
No significant difference (p$>$0.05) was found in the VIBES modality compared to the without VIBES modality in all experimental outcome measures ( i.e. number of fallen cylinders, reaction time and slip). 

\textbf{Prosthetic User} - Table \ref{tab:pu_slip} reports experimental results of the porsthetic user participant.
\begin{table}[]
\small
\MFUnocap{for}%
\MFUnocap{the}%
\MFUnocap{of}%
\MFUnocap{and}%
\MFUnocap{in}%
\MFUnocap{from}%
\caption{\capitalisewords{Pilot Experiments - Slippage Experiment: Reaction Time, Number of fallen cylinder and Slip}}
\label{tab:pu_slip}
\resizebox{\columnwidth}{!}{%
\begin{tabular}{cccc|ccc}
\multicolumn{4}{c|}{\textbf{WITH VIBES}} & \multicolumn{3}{c}{\textbf{WITHOUT VIBES}} \\ 
\multicolumn{1}{l}{Trials} &
  \begin{tabular}[c]{@{}c@{}}Reaction \\Time [s]\end{tabular} &
  Fallen Cylinder &
  Slip [cm] &
  \begin{tabular}[c]{@{}c@{}}Reaction \\Time [s]\end{tabular} &
  Fallen Cylinder &
  Slip [cm] \\ \hline
1       & 0.28      & 0      & 1.2       & 0.36           & 0          & 2.5          \\
2       & 0.33      & 0      & 5.1       & 0.28           & 0          & 1.6          \\
3       & 0.37      & 1      & 11.5      & 0.31           & 0          & 2.1          \\
4       & 0.32      & 0      & 3.5       & 0.33           & 0          & 3.5          \\
5       & 0.28      & 0      & 1.6       & 0.40           & 1          & 8.2         
\end{tabular}%
}
\vspace{-0.3cm}
\end{table}
The number of fallen cylinders was the same in both modalities. 
The medium slip with the VIBES was 4.6 cm, while it was 3.58 cm without.
The average reaction time was lower with the VIBES (0.31 s) compared to without it (0.34 s).
\subsubsection{Questionnaires}
\begin{table}[]
  \small
  \MFUnocap{for}%
  \MFUnocap{the}%
  \MFUnocap{of}%
  \MFUnocap{and}%
  \MFUnocap{in}%
  \MFUnocap{from}%
  \caption{\capitalisewords{System Validation and Pilot Experiments - NASA RTLX index for each experiment.}}
  \label{tab:NASA}
  \resizebox{\columnwidth}{!}{%
\begin{tabular}{c|cccc}
\multirow{2}{*}{\textbf{Experiment}} &
  \multicolumn{4}{c}{\textbf{NASA RTLX Index}} \\
 &
  \multicolumn{2}{c|}{\textbf{Able-Bodied Subjects}} &
  \multicolumn{2}{c}{\textbf{Prosthetic User}} \\
\textbf{} &
  \begin{tabular}[c]{@{}c@{}}Without\\  the VIBES\end{tabular} &
  \multicolumn{1}{c|}{\begin{tabular}[c]{@{}c@{}}With \\ the VIBES\end{tabular}} &
  \begin{tabular}[c]{@{}c@{}}Without \\ the VIBES\end{tabular} &
  \begin{tabular}[c]{@{}c@{}}With \\ the VIBES\end{tabular} \\ \hline
\begin{tabular}[c]{@{}c@{}}Active Texture \\ Identification\end{tabular} &
  46.3 &
  \multicolumn{1}{c|}{45.7} &
  18.3 &
  35.0 \\
Fragile Object &
  43.9 &
  \multicolumn{1}{c|}{42.4} &
  26.7 &
  33.3 \\
Slippage &
  61.4 &
  \multicolumn{1}{c|}{58.8} &
  25.0 &
  30.0
\end{tabular}%
  }
  \vspace{-0.6cm}
  \end{table}
\textbf{Able-bodied} - Results of the NASA RTLX workload assessment are reported in Table \ref{tab:NASA}.
The NASA RTLX scores to asses the overall workload were lower with the VIBES compared to without it in all validation experiments.
The Lilliefors test confirmed that the NASA RTLX scores were normally distributed. Additionally, the paired two-sample t-test showed no significant differences between the VIBES and non-VIBES conditions in all experiments (p$>$0.05).
For a detailed report on the NASA RTLX results for each question across the three experiments with able-bodied participants, please refer to Fig. S1 in the Supplemental Materials. Overall, in the three experiments, mental demand was higher or equal with the VIBES compared to without it, while physical and temporal demands were lower or equal with the VIBES. Performance was perceived as better without VIBES in the Active Texture Identification and Slippage Experiments, but worse without the VIBES in the Fragile Object Experiment. Effort was consistently higher with VIBES across all experiments. Frustration was lower with the VIBES in the Active Texture Identification and Fragile Object Experiments, but higher in the Slippage Experiment
Table \ref{tab:likert} shows the results of the qualitative questionnaires with the mean and the standard deviation of the scores of each question for the able-bodied subjects. 
\begin{table*}[]
\MFUnocap{for}%
\MFUnocap{the}%
\MFUnocap{of}%
\MFUnocap{and}%
\MFUnocap{in}%
\MFUnocap{from}%
\caption{\capitalisewords{System Validation Experiments - Questionnaires: Results of the qualitative questionnaires evaluated on a 7-point Likert scale (1: Strongly disagree, 7: Strongly agree)}}
\label{tab:likert}
\resizebox{\textwidth}{!}{%
\begin{tabular}{l|ll|c}
\hline
\multicolumn{1}{c|}{\multirow{2}{*}{\textbf{Questions}}}                                                                                  & \multicolumn{2}{c|}{\textbf{Able-Bodied}} & \multicolumn{1}{c}{\textbf{Prosthetic User}} \\
\multicolumn{1}{c|}{}                                                                                                                     & \textbf{Mean}     & \textbf{Std. Dev}     & \textbf{Score}                               \\ \midrule
1.   It was easy to wear and use the VIBES device.                                                                        & 5
& 1.0& 7                                            \\ \midrule
2.    I was feeling uncomfortable using the VIBES.                                                                  &2.6
& 1.0& 3\\ \midrule
3.    I was well-isolated from the external noises during the experiments.                                                                & 6.4
& 0.6& 7                                            \\ \midrule
4.   I was able to hear the sounds made by the actuator of the device.                                                  & 1.5
& 0.5
& 1                                            \\ \midrule
5.   The stimuli provided by the cutaneous device allowed for the detection of slippage.                                                                      &2.6
& 1.2
& 4\\ \midrule
6.    The stimuli provided by the cutaneous device did not allow to detect slippage.
        & 5.3
& 1.2& 4\\ \midrule
7.   The stimuli provided by the cutaneous device allowed to discriminate different texture levels.                                               &3.5
& 1.4& 4                                            \\ \midrule
8.   The stimuli provided by the cutaneous device did not allow for discrimination of different texture levels.                                                 & 4.3
& 1.6& 4\\ \midrule
9. The stimuli provided by the cutaneous device allowed to improve the grasp efficacy.                                                           & 3.4
& 0.6& 2\\ \midrule
10. The stimuli provided by the cutaneous device did not allow to improve the grasp efficacy.                                                & 4.5
& 1.1& 5\\ \midrule
11.    I felt hampered by the vibrotactile stimuli.                                                          & 1.6
&0.4& 4\\ \midrule
12.   I was able to see the device during the experiment.                      & 1
& 0
& 1                                            \\ \midrule
13.    At the end of the experiment, I felt tired.                                                   & 2.2
& 1.2& 5\\ \midrule
14.   I prefer to receive the stimuli at two different body locations, e.g. A and B conditions.                  & 2.7
& 1.6& 4\\ \midrule
15.    I prefer to receive the stimuli at the same body location, e.g. A condition.
                                                    & 3.9& 1.6& 5\\ \midrule
16.   I think I would have done the experiments better without the VIBES.       & 3.5& 1.8& 4\\ \bottomrule
\end{tabular}
}
\end{table*}

\textbf{Prosthetic User} - About the prosthetic user, the SUS questionnaire resulted in a positive score of 77.5 (average SUS score of 68 at the 50$^{th}$ percentile).

Results of the Qualitative Questionnaires evaluated on 7-point Likert scale are reported in the last column of Table \ref{tab:likert}.
The VIBES resulted easy to wear and use, and the subject did not perceive any improvements or deterioration of the performance with the VIBES in discriminating textures or detecting slippage. 
However, the subject perceived that the VIBES did not allow for improved grasp efficacy.
The subject preferred to receive stimuli at the same body location (e.g. A condition).
Results of the NASA are reported in Table \ref{tab:NASA} and showed a slight increase in the workload when the VIBES was active.
\subsubsection{Rubber Hand Illusion Experiment}
\begin{figure*}
\centering
\vspace{-0.3cm}
\includegraphics[width=0.75\textwidth]{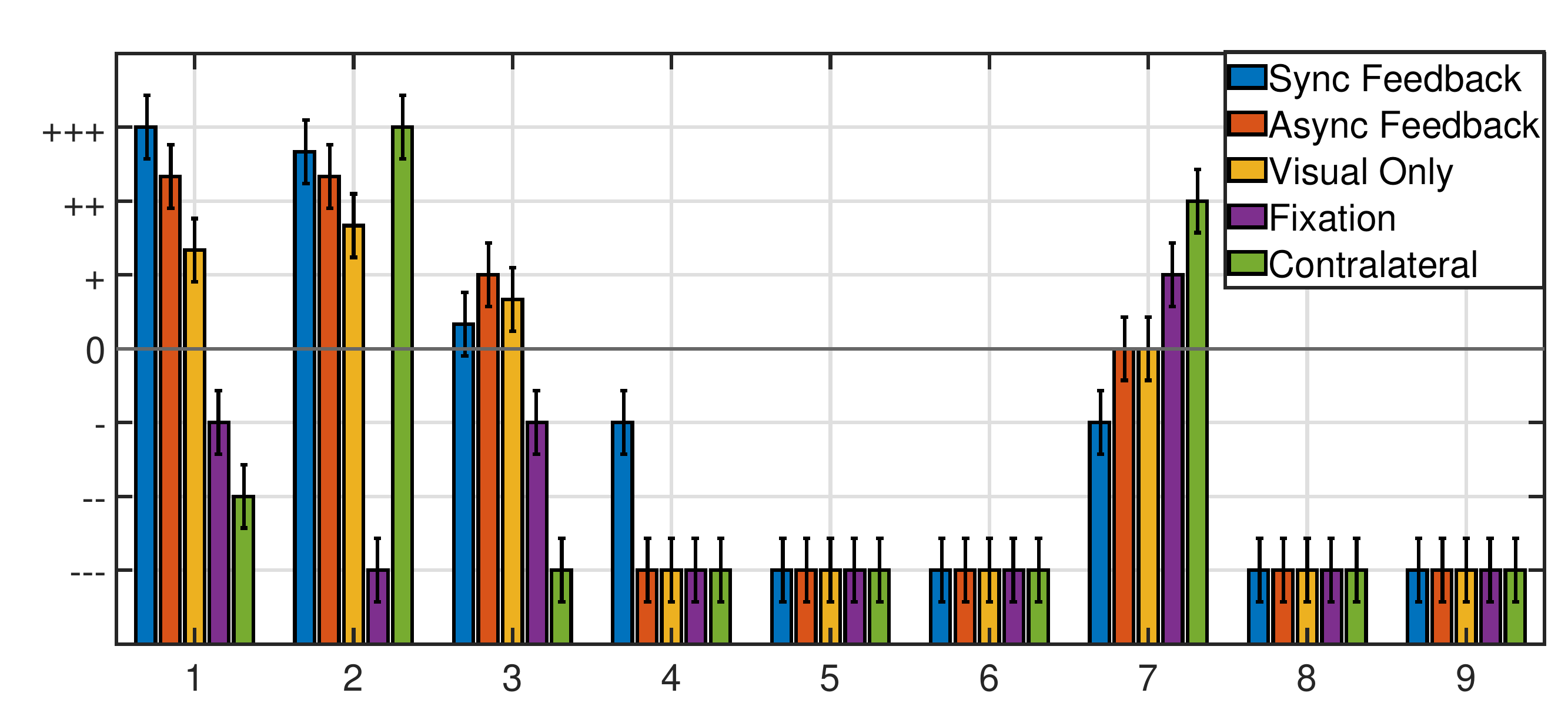}
\caption{RHI: Results of the qualitative questionnaire for the five experimental conditions; Error bars indicate 95\% CI ($\pm$0.43) from a multiple comparisons procedure. Significance judged by non-overlap of CI. Statements 1–3 = predicted phenomena (please refer to Table \ref{tab:quest}). Statements 4–9 = controls. Each of the nine statements was assessed using a seven-point visual scale, which spans from $'$strongly disagree$'$  $(---)$ to $'$strongly agree$'$ ($+++$).}
\label{fig:RHI}
\vspace{-0.6cm}
\end{figure*}

 \textbf{Prosthetic User} - Results of the RHI subjective questionnaire are reported in Fig. \ref{fig:RHI}. 
Significance was judged by non-overlap of credible intervals.
All the illusion statements for the Synchronous Feedback and Asynchronous Feedback experimental conditions were significantly higher than the control statements. 
No significant differences exist between the Synchronous and Asynchronous conditions (with overlapping credibility intervals), except for control statements 4 and 7. 
In the Visual Only condition, the illusion statements were higher than the control statements except for statement number 7. 
In the illusion statements number 1 and 2, the VO condition was significantly different from the Synchronous Feedback condition. 
No significant difference was found between VO and AF in statements number 2 and 3, while AF was significantly different from VO in the first statement.
In the CL condition, the subject did not experience ownership, except for statement number 2, even if the RHI was administered with the traditional protocol.
\section{Discussion}
Overall, our study seems to reveal an improvement in texture perception achieved through the VIBES feedback without causing discomfort.
It remains neutral in influencing slip detection or fragile object movements. 
The notable increase in embodiment warrants further investigation with a larger subject pool.
\vspace{-0.4cm}
\subsection{Psychophysical Characterization}
The System Characterization of the VIBES revealed consistent behaviours among able-bodied subjects when discriminating roughness stimuli on the forearm with actuators positioned on the A and the B experimental conditions.
The JND was 58.79 $\mu$m in the A experimental condition and 64.10 $\mu$m in the B experimental condition.
Perception was not affected by the position of the actuator on the forearm skin. 
It should be noted that the JND results are in accordance with the results obtained by passive exploration of texture on the index fingertip (JND of 87.30 $\mu$m), reported in Section III and described in our previous work \cite{ivani2023vibes}.
Various subjects exhibit different sensitivities to stimuli, yet the behaviour remains consistent within each condition. Despite these differences, the total Just Noticeable Difference (JND) in each condition is consistently low, indicating that subjects can effectively distinguish between stimuli.

About the Pilot Experiments, the Psychophysical Characterization Experiment with a prosthetic user produced positive results, revealing a Just Noticeable Difference of 44.07$\mu$m, lower than the one for able-bodied subjects.
\vspace{-0.4cm}
\subsection{Active Texture Identification Experiment}
The System Validation of the VIBES with able-bodied participants revealed significant improvements in the performance in the Active Texture Identification Experiment with an accuracy of 62\% in recognizing the matching sandpaper with the VIBES compared to 50\% without the VIBES.
Thus, the subjects were able to recognize the texture better with the VIBES than without it.
The prosthetic user participant could also better identify the sandpaper presented with the VIBES feedback than without it in the Active Texture Identification Experiment.
Indeed, the accuracy with the vibrotactile feedback was 52\% compared to 40\% without the feedback.
This outcome emphasizes that the 10 subjects effectively utilized and comprehended the device's primary function of transmitting texture information.
We acknowledge the exclusion of two participants, resulting in a reduction of the sample size to 10, representing a 17\% exclusion rate. 
Nevertheless, the positive outcomes observed in the study remain encouraging, although future investigations are needed to further strengthen the conclusions of our work. Subsequent investigations will delve deeper into this aspect, with efforts to increase the number of participants.
Indeed, the variation in skin innervation between the forearm and hand could account for the presence of two outliers between able-bodied participants.
These two subjects identify texture stimuli on the index finger but exhibit very low sensitivity on the forearm.
Also, it's worth mentioning that the active texture exploration in this experiment involved muscle contraction (including raising the handle), potentially decreasing the sensitivity of the subjects compared to the Psychophysical Characterization Experiment \cite{ivani2023vibes}. 
As a result, we opted to exclude the two participants from the experiment.
Future investigations will further explore this aspect, with efforts to increase the number of participants.

The choice of sandpapers remained consistent with the System Characterization Experiment. 
They were specifically selected to align with the method of constant stimuli, ensuring that the reference stimulus (texture 3) is easily distinguishable from stimulus 1 and stimulus 5. 
Consequently, there may be notable distinctions in identifying texture 1 and texture 2 compared to texture 5.
While the experiments present promising results, attaining 100\% accuracy proved unfeasible. 
It's worth noticing that even able-bodied individuals may not achieve 100\% accuracy in discriminating similar roughness using their own fingers (Motamedi et al. ~\cite{motamedi2016use}).
Note that in the Active Texture Identification experiments, both prosthetic users and able-bodied individuals demonstrated an ability to identify matching sandpaper textures without feedback, surpassing chance levels. 
This outcome might be attributed to the propagation of intrinsic and extrinsic feedback in recognizing different textures (Amoruso et al.~\cite{amoruso2022intrinsic}).
Although the SoftHand Pro incorporates damping elements, it is conceivable that the transmission of vibrations through the socket or the handle may have influenced participants' perception. 
Further investigations are needed to explore this aspect in more detail.
\vspace{-0.4cm}
\subsection{Fragile Object Experiment and Slippage Experiment}
In the Fragile Object Experiment and Slippage Experiment with able-bodied subjects, no significant differences were found in the two experimental conditions (with the VIBES and without the VIBES).
Improved performances in terms of times to perform the tasks and number of regrips were found in the Pilot Fragile Object Experiment with the VIBES.
However, an increase in broken fuses (from 3 to 6) was also noticed.
About the Slippage Experiment, the same number of cylinders slipped with the VIBES and without it, but the reaction time and the slip were lower with the VIBES compared to without it.
Based on these findings, it seems that VIBES had neither a positive nor negative impact on the dexterity and manual usability of the prosthesis user, accordingly to Raveh et al. results~\cite{raveh2018myoelectric}.
While the results of the tasks yielded neutral outcomes, it is crucial to underscore that the primary objective of these tasks was to assess the effects of the device in diverse scenarios beyond texture recognition, with the intent of identifying potential positive or negative implications. Consequently, no adverse effects were observed on either performance or user experience, indicating that the device had no discernible influence in these particular scenarios.
In these experiments, alternative feedback may be used for slip detection and delicate object manipulation.
Paired with our WISH device, a soft pneumatic system delivering contact and grip force cues, the VIBES can transform the SHP into a versatile prosthesis with dual haptic feedback options (Barontini et al. \cite{barontini2021wearable}).
\vspace{-0.5cm}
\subsection{Questionnaires}
Based on prosthetic user feedback, the VIBES provided an intuitive experience, supported by a high SUS questionnaire score, a promising sign of usability and satisfaction.
Nevertheless, the user did not discern any notable differences in texture discrimination or slip detection performance. 
Additionally, the user perceived that the VIBES did not contribute to the improvement of grasp efficacy.

Regarding the able-bodied participants, despite the performance improvement shown in Fig. \ref{fig:active}, the lack of reported awareness of this improvement in the results of the Qualitative Questionnaire (Table \ref{tab:likert}) prompts further investigation.
One possible explanation is that a non-intrusive stimulus allows subjects to concentrate solely on the task without distractions. Consequently, they may unknowingly integrate the stimulus information into the discrimination process. Alternatively, individual differences in sensory perception thresholds may contribute to variations in the perception of the stimulus enhancement among subjects. Additionally, factors such as cognitive load, attentional allocation, and task demands could affect subjects' ability to detect subtle changes in sensory stimuli.
However, these explanations are speculative, and further investigations are needed to clarify the underlying mechanisms and validate these hypotheses.
Furthermore, able-bodied participants reported that the device did not aid in detecting slippage, and there were no significant differences in grasp efficacy performance.

Regarding workload, there was a slight decrease for able-bodied individuals and an increase for prosthetic users.
Among the able-bodied participants, the task with a higher workload was the Slippage experiment, whereas, for the prosthetic user, it was the Fragile Object Experiment.
Given the comparable NASA RTLX and NASA-TLX indices, all experiments for able-bodied participants showed workloads similar to driving activities (Byers et al. \cite{Byers1989TraditionalAR}, Grier et al. \cite{doi:10.1177/1541931215591373}.
The prosthetic user's workload was lower and comparable to routine activities such as conversation, telephone inquiries, and use of home medical devices \cite{doi:10.1177/1541931215591373}. 
Hence, the VIBES doesn't require the user's full attention, underscoring the system's intuitiveness.
However, the subjective nature of the results underscores the need for additional testing.
\vspace{-0.3cm}
\subsection{Rubber Hand Illusion Experiment}
In the evaluation of SHP embodiment within the Rubber Hand Illusion Experiment, a substantial enhancement in embodiment, notably in the ownership domain, was observed with the inclusion of the VIBES compared to its absence. This significant increase is consistent with findings reported in previous studies \cite{marasco2011robotic,6856175}.
These results underscore the positive implications of integrating the VIBES into the SHP, especially considering the pivotal role of embodiment in addressing prosthetic abandonment rates.
Notably, the prosthetic side demonstrated superior performance compared to the contralateral side, possibly influenced by the participant's limb agenesis.
Despite the adherence to an asynchrony time that reduced the illusionary experience in past studies \cite{Kokkinara2014MeasuringTE,10.1371/journal.pone.0087013,articlegrasp}, in the Asynchrony feedback condition responses indicated that the illusion persisted. 
These outcomes emphasize the necessity for future studies involving a larger sample size.
Such investigations could offer more profound insights into the effects of feedback on the SHP embodiment. 
\section{Conclusion}\label{sec:5}
This study characterizes, validates, and tests the VIBES device - a vibrotactile integrated feedback system designed to convey texture and contact cues to prosthetic users.
In comparison to our previous study \cite{ivani2023vibes}, the VIBES was here characterized in an experiment with 10 able-bodied subjects.
Positive outcomes in tactile cue discrimination were observed across two experimental conditions (referred to as A and B) designed to test actuator positioning.
Moreover, active texture identification experiments involving a prosthetic user and able-bodied participants indicate that VIBES effectively transmits texture feedback in active tasks.
 The study investigated the effects of VIBES beyond texture recognition through Fragile Object and Slippage experiments. Despite yielding neutral results, no noticeable improvements in performance or user experience were detected. Additionally, questionnaire responses from a prosthetic user indicated no discernible differences in texture discrimination or slip detection performance, nor any perceived improvement in grasp efficacy. Regarding workload, there was a slight decrease observed for able-bodied individuals, contrasting with an increase for prosthetic users. Nevertheless, the system's workload remained comparable to routine activities, suggesting its intuitiveness. Notably, prosthetic user embodiment increased with VIBES in a Rubber Hand Illusion experiment. Further research is essential to comprehensively evaluate the device's effectiveness and potential.
\section*{ACKNOWLEDGMENT}
The authors thank Marina Gnocco, Mattia Poggiani, and Manuel Barbarossa for their support in this work.
\bibliographystyle{IEEEtran}
\bibliography{mybibfile}

\begin{thebibliography}{10}
\providecommand{\url}[1]{#1}
\csname url@rmstyle\endcsname
\providecommand{\newblock}{\relax}
\providecommand{\bibinfo}[2]{#2}
\providecommand\BIBentrySTDinterwordspacing{\spaceskip=0pt\relax}
\providecommand\BIBentryALTinterwordstretchfactor{4}
\providecommand\BIBentryALTinterwordspacing{\spaceskip=\fontdimen2\font plus
\BIBentryALTinterwordstretchfactor\fontdimen3\font minus
  \fontdimen4\font\relax}
\providecommand\BIBforeignlanguage[2]{{%
\expandafter\ifx\csname l@#1\endcsname\relax
\typeout{** WARNING: IEEEtran.bst: No hyphenation pattern has been}%
\typeout{** loaded for the language `#1'. Using the pattern for}%
\typeout{** the default language instead.}%
\else
\language=\csname l@#1\endcsname
\fi
#2}}

\bibitem{choi2012vibrotactile}
S.~Choi and K.~J. Kuchenbecker, ``Vibrotactile display: Perception, technology,
  and applications,'' \emph{Proceedings of the IEEE}, vol. 101, no.~9, pp.
  2093--2104, 2012.

\bibitem{thomas2019comparison}
N.~Thomas, G.~Ung, C.~McGarvey, and J.~D. Brown, ``Comparison of vibrotactile
  and joint-torque feedback in a myoelectric upper-limb prosthesis,''
  \emph{Journal of NeuroEngineering and Rehabilitation}, vol.~16, no.~1, pp.
  1--18, 2019.

\bibitem{akhtar2020touch}
A.~Akhtar, J.~Cornman, J.~Austin, and D.~Bala, ``Touch feedback and contact
  reflexes using the psyonic ability hand,'' 2020.

\bibitem{kim2012haptic}
K.~Kim and J.~E. Colgate, ``Haptic feedback enhances grip force control of
  semg-controlled prosthetic hands in targeted reinnervation amputees,''
  \emph{IEEE Transactions on Neural Systems and Rehabilitation Engineering},
  vol.~20, no.~6, pp. 798--805, 2012.

\bibitem{antfolk2012sensory}
C.~Antfolk, A.~Bj{\"o}rkman, S.-O. Frank, F.~Sebelius, G.~Lundborg, and
  B.~Rosen, ``Sensory feedback from a prosthetic hand based on air-mediated
  pressure from the hand to the forearm skin,'' \emph{Journal of rehabilitation
  medicine}, vol.~44, no.~8, p. 702, 2012.

\bibitem{visell}
\BIBentryALTinterwordspacing
Y.~Shao, V.~Hayward, and Y.~Visell, ``Compression of dynamic tactile
  information in the human hand,'' \emph{Science Advances}, vol.~6, no.~16, p.
  eaaz1158, 2020. [Online]. Available:
  \url{https://www.science.org/doi/abs/10.1126/sciadv.aaz1158}
\BIBentrySTDinterwordspacing

\bibitem{actionsomatosensory}
\BIBentryALTinterwordspacing
K.~E.R., S.~J.H., J.~T.M., S.~S.A., H.~A.J., and M.~S., \emph{The Somatosensory
  System: Receptors and Central Pathways}.\hskip 1em plus 0.5em minus
  0.4em\relax New York, NY: McGraw-Hill Education, 2014. [Online]. Available:
  \url{neurology.mhmedical.com/content.aspx?aid=1101679414}
\BIBentrySTDinterwordspacing

\bibitem{sensinger2020review}
J.~W. Sensinger and S.~Dosen, ``A review of sensory feedback in upper-limb
  prostheses from the perspective of human motor control,'' \emph{Frontiers in
  Neuroscience}, vol.~14, p. 345, 2020.

\bibitem{raveh2018myoelectric}
E.~Raveh, S.~Portnoy, and J.~Friedman, ``Myoelectric prosthesis users improve
  performance time and accuracy using vibrotactile feedback when visual
  feedback is disturbed,'' \emph{Archives of physical medicine and
  rehabilitation}, vol.~99, no.~11, pp. 2263--2270, 2018.

\bibitem{markovic2018clinical}
M.~Markovic, M.~A. Schweisfurth, L.~F. Engels, T.~Bentz, D.~W{\"u}stefeld,
  D.~Farina, and S.~Dosen, ``The clinical relevance of advanced artificial
  feedback in the control of a multi-functional myoelectric prosthesis,''
  \emph{Journal of neuroengineering and rehabilitation}, vol.~15, no.~1, pp.
  1--15, 2018.

\bibitem{segil2022measuring}
J.~L. Segil, L.~M. Roldan, and E.~L. Graczyk, ``Measuring embodiment: A review
  of methods for prosthetic devices,'' \emph{Frontiers in Neurorobotics},
  vol.~16, p. 902162, 2022.

\bibitem{zbinden2022prosthetic}
J.~Zbinden, E.~Lendaro, and M.~Ortiz-Catalan, ``Prosthetic embodiment:
  systematic review on definitions, measures, and experimental paradigms,''
  \emph{Journal of NeuroEngineering and Rehabilitation}, vol.~19, no.~1, p.~37,
  2022.

\bibitem{doi:10.3109/02844310903113107}
B.~Rosen, H.~Ehrsson, C.~Antfolk, C.~Cipriani, F.~Sebelius, and G.~Lundborg,
  ``Referral of sensation to an advanced humanoid robotic hand prosthesis,''
  \emph{Scandinavian journal of plastic and reconstructive surgery and hand
  surgery / Nordisk plastikkirurgisk forening [and] Nordisk klubb for
  handkirurgi}, vol.~43, pp. 260--6, 01 2009.

\bibitem{6856175}
M.~D'Alonzo, F.~Clemente, and C.~Cipriani, ``Vibrotactile stimulation promotes
  embodiment of an alien hand in amputees with phantom sensations,'' \emph{IEEE
  Transactions on Neural Systems and Rehabilitation Engineering}, vol.~23,
  no.~3, pp. 450--457, 2015.

\bibitem{marasco2011robotic}
P.~D. Marasco, K.~Kim, J.~E. Colgate, M.~A. Peshkin, and T.~A. Kuiken,
  ``Robotic touch shifts perception of embodiment to a prosthesis in targeted
  reinnervation amputees,'' \emph{Brain}, vol. 134, no.~3, pp. 747--758, 2011.

\bibitem{ivani2023vibes}
A.~S. Ivani, F.~Barontini, M.~G. Catalano, G.~Grioli, M.~Bianchi, and
  A.~Bicchi, ``Vibes: Vibro-inertial bionic enhancement system in a prosthetic
  socket,'' 2023, pp. 1--6.

\bibitem{godfrey2018softhand}
S.~B. Godfrey, K.~D. Zhao, A.~Theuer, M.~G. Catalano, M.~Bianchi, R.~Breighner,
  D.~Bhaskaran, R.~Lennon, G.~Grioli, M.~Santello, \emph{et~al.}, ``The
  softhand pro: Functional evaluation of a novel, flexible, and robust
  myoelectric prosthesis,'' \emph{PloS one}, vol.~13, no.~10, p. e0205653,
  2018.

\bibitem{amoruso2022intrinsic}
E.~Amoruso, L.~Dowdall, M.~Kollamkulam, O.~Ukaegbu, P.~Kieliba, T.~Ng,
  H.~Dempsey-Jones, D.~Clode, and T.~Makin, ``Intrinsic somatosensory feedback
  supports motor control and learning to operate artificial body parts,''
  \emph{Journal of Neural Engineering}, vol.~19, no.~1, p. 016006, 2022.

\bibitem{NMMI}
C.~Della~Santina, C.~Piazza, G.~Gasparri, M.~Bonilla, M.~Catalano, G.~Grioli,
  M.~Garabini, and A.~Bicchi, ``The quest for natural machine motion: An open
  platform to fast-prototyping articulated soft robots,'' \emph{IEEE Robotics
  and Automation Magazine}, vol.~PP, pp. 1--1, 02 2017.

\bibitem{paperfani}
S.~Fani, K.~D. Blasio, M.~Bianchi, M.~G. Catalano, G.~Grioli, and A.~Bicchi,
  ``Relaying the high-frequency contents of tactile feedback to robotic
  prosthesis users: Design, filtering, implementation, and validation,''
  \emph{IEEE Robotics and Automation Letters}, vol.~4, no.~2, pp. 926--933,
  2019.

\bibitem{landin2010dimensional}
N.~Landin, J.~M. Romano, W.~McMahan, and K.~J. Kuchenbecker, ``Dimensional
  reduction of high-frequency accelerations for haptic rendering.''\hskip 1em
  plus 0.5em minus 0.4em\relax Springer, 2010.

\bibitem{9773011}
H.~Lee, G.~I. Tombak, G.~Park, and K.~J. Kuchenbecker, ``Perceptual space of
  algorithms for three-to-one dimensional reduction of realistic vibrations,''
  \emph{IEEE Transactions on Haptics}, vol.~15, no.~3, pp. 521--534, 2022.

\bibitem{jones2012application}
L.~A. Jones and H.~Z. Tan, ``Application of psychophysical techniques to haptic
  research,'' \emph{IEEE transactions on haptics}, vol.~6, no.~3, pp. 268--284,
  2012.

\bibitem{libouton2010tactile}
X.~Libouton, O.~Barbier, L.~Plaghki, and J.-L. Thonnard, ``Tactile roughness
  discrimination threshold is unrelated to tactile spatial acuity,''
  \emph{Behavioural brain research}, vol. 208, no.~2, pp. 473--478, 2010.

\bibitem{boundy2017speed}
Z.~M. Boundy-Singer, H.~P. Saal, and S.~J. Bensmaia, ``Speed invariance of
  tactile texture perception,'' \emph{Journal of neurophysiology}, vol. 118,
  no.~4, pp. 2371--2377, 2017.

\bibitem{SAAL201899}
\BIBentryALTinterwordspacing
H.~P. Saal, A.~K. Suresh, L.~E. Solorzano, A.~I. Weber, and S.~J. Bensmaia,
  ``The effect of contact force on the responses of tactile nerve fibers to
  scanned textures,'' \emph{Neuroscience}, vol. 389, pp. 99--103, 2018, sensory
  Sequence Processing in the Brain. [Online]. Available:
  \url{https://www.sciencedirect.com/science/article/pii/S0306452217305900}
\BIBentrySTDinterwordspacing

\bibitem{10.1167/12.11.26}
\BIBentryALTinterwordspacing
A.~Moscatelli, M.~Mezzetti, and F.~Lacquaniti, ``{Modeling psychophysical data
  at the population-level: The generalized linear mixed model},'' \emph{Journal
  of Vision}, vol.~12, no.~11, pp. 26--26, 10 2012. [Online]. Available:
  \url{https://doi.org/10.1167/12.11.26}
\BIBentrySTDinterwordspacing

\bibitem{Georgsson2020NASARA}
\BIBentryALTinterwordspacing
M.~Georgsson, ``Nasa rtlx as a novel assessment tool for determining cognitive
  load and user acceptance of expert and user-based usabilityevaluation
  methods,'' 2020. [Online]. Available:
  \url{https://api.semanticscholar.org/CorpusID:222448058}
\BIBentrySTDinterwordspacing

\bibitem{Lewis2018TheSU}
J.~R. Lewis, ``The system usability scale: Past, present, and future,''
  \emph{International Journal of Human–Computer Interaction}, vol.~34, pp.
  577 -- 590, 2018.

\bibitem{barontini2021wearable}
F.~Barontini, M.~G. Catalano, G.~Grioli, M.~Bianchi, and A.~Bicchi, ``Wearable
  integrated soft haptics in a prosthetic socket,'' \emph{IEEE Robotics and
  Automation Letters}, vol.~6, no.~2, pp. 1785--1792, 2021.

\bibitem{Engels2018DigitalEW}
L.~F. Engels, L.~Cappello, and C.~Cipriani, ``Digital extensions with bi-axial
  fingertip sensors for supplementary tactile feedback studies,'' \emph{2018
  7th IEEE International Conference on Biomedical Robotics and Biomechatronics
  (Biorob)}, pp. 1199--1204, 2018.

\bibitem{grandini2020metrics}
M.~Grandini, E.~Bagli, and G.~Visani, ``Metrics for multi-class classification:
  an overview,'' \emph{arXiv preprint arXiv:2008.05756}, 2020.

\bibitem{Kokkinara2014MeasuringTE}
\BIBentryALTinterwordspacing
E.~Kokkinara and M.~Slater, ``Measuring the effects through time of the
  influence of visuomotor and visuotactile synchronous stimulation on a virtual
  body ownership illusion,'' \emph{Perception}, vol.~43, pp. 43 -- 58, 2014.
  [Online]. Available: \url{https://api.semanticscholar.org/CorpusID:12473599}
\BIBentrySTDinterwordspacing

\bibitem{10.1371/journal.pone.0087013}
\BIBentryALTinterwordspacing
R.~Bekrater-Bodmann, J.~Foell, M.~Diers, S.~Kamping, M.~Rance, P.~Kirsch,
  J.~Trojan, X.~Fuchs, F.~Bach, H.~K. Çakmak, H.~Maaß, and H.~Flor, ``The
  importance of synchrony and temporal order of visual and tactile input for
  illusory limb ownership experiences – an fmri study applying virtual
  reality,'' \emph{PLOS ONE}, vol.~9, pp. 1--10, 01 2014. [Online]. Available:
  \url{https://doi.org/10.1371/journal.pone.0087013}
\BIBentrySTDinterwordspacing

\bibitem{articlegrasp}
T.~Huynh, R.~Bekrater-Bodmann, J.~Fröhner, J.~Vogt, and P.~Beckerle, ``Robotic
  hand illusion with tactile feedback: Unravelling the relative contribution of
  visuotactile and visuomotor input to the representation of body parts in
  space,'' \emph{PLoS ONE}, vol.~14, p. e0210058, 01 2019.

\bibitem{motamedi2016use}
M.~R. Motamedi, J.-P. Roberge, and V.~Duchaine, ``The use of vibrotactile
  feedback to restore texture recognition capabilities, and the effect of
  subject training,'' \emph{IEEE Transactions on Neural Systems and
  Rehabilitation Engineering}, vol.~25, no.~8, pp. 1230--1239, 2016.

\bibitem{Byers1989TraditionalAR}
\BIBentryALTinterwordspacing
J.~C. Byers, A.~C. Bittner, and S.~G. Hill, ``Traditional and raw task load
  index (tlx) correlations: Are paired comparisons necessary? in a,'' 1989.
  [Online]. Available: \url{https://api.semanticscholar.org/CorpusID:58504578}
\BIBentrySTDinterwordspacing

\bibitem{doi:10.1177/1541931215591373}
R.~A. Grier, ``How high is high? a meta-analysis of nasa-tlx global workload
  scores,'' \emph{Proceedings of the Human Factors and Ergonomics Society
  Annual Meeting}, vol.~59, no.~1, pp. 1727--1731, 2015.

\end{thebibliography}
\end{document}